%% file: self6dpp.tex
\documentclass[10pt,journal,compsoc,twoside]{IEEEtran}
\input{packages}
\input{macros}

\newcommand\copyrighttext{%
  \footnotesize \textcopyright 2021 IEEE. Personal use of this material is permitted.  
  Permission from IEEE must be obtained for all other uses, in any current or future media, including reprinting/republishing this material for advertising or promotional purposes, creating new collective works, for resale or redistribution to servers or lists, or reuse of any copyrighted component of this work in other works.
  DOI: \href{https://doi.org/10.1109/TPAMI.2021.3136301}{10.1109/TPAMI.2021.3136301}}
\newcommand\copyrightnotice{%
\begin{tikzpicture}[remember picture,overlay]
\node[anchor=south,yshift=5pt] at (current page.south)
{\fbox{\parbox{\dimexpr\textwidth-\fboxsep-\fboxrule\relax}{\copyrighttext}}};
\end{tikzpicture}%
}

\begin{document}
\title{Occlusion-Aware Self-Supervised Monocular 6D Object Pose Estimation}

\author{Gu~Wang$^\dagger$\,\textsuperscript{\orcidicon{0000-0002-0759-0782}}, %~\IEEEmembership{Student Member,~IEEE,}
        Fabian~Manhardt$^\dagger$\,\textsuperscript{\orcidicon{0000-0002-4577-4590}},
        %~\IEEEmembership{Student Member,~IEEE,}\\
        Xingyu~Liu\,\textsuperscript{\orcidicon{0000-0003-1156-2263}}, %~\IEEEmembership{Student Member,~IEEE,}
        % Jianzhun~Shao, %~\IEEEmembership{Student Member,~IEEE,}
        Xiangyang~Ji\corrAuthor\,\textsuperscript{\orcidicon{0000-0002-7333-9975}},
        % Nassir~Navab,
        and~Federico~Tombari\,\textsuperscript{\orcidicon{0000-0001-5598-5212}}%,~\IEEEmembership{Fellow,~IEEE}% <-this % stops a space
    \IEEEcompsocitemizethanks{
        \IEEEcompsocthanksitem Gu Wang, Xingyu Liu and Xiangyang Ji are with the Department of Automation, Tsinghua University, Beijing 100084, China, and also with BNRist, Beijing 100084, China.
        E-mail: \{wangg16, liuxy21\}@mails.tsinghua.edu.cn, xyji@tsinghua.edu.cn.
        \IEEEcompsocthanksitem Fabian Manhardt is with Google Inc., 8002 Zurich, Switzerland.
        \protect\\
        E-mail: fabianmanhardt@google.com. %\protect\\
        \IEEEcompsocthanksitem Federico Tombari is with Google Inc., 8002 Zurich, Switzerland, and also with the Technical University of Munich, 80333 M\"unchen, Germany. E-mail: tombari@in.tum.de.
    }% <-this % stops an unwanted space
    \thanks{Manuscript received 1 May 2021; revised 6 Oct. 2021; accepted 14 Dec. 2021.
        \protect\\
        Date of publication 0 . 0000; date of current version 0 . 0000.
        \protect\\
        This work was supported in part by the National Key R\&D Program of China under Grant 2018AAA0102801, 
        in part by China Scholarship Council (CSC) Grant \#201906210393, 
        and in part by National Natural Science Foundation of China under Grant 61620106005. 
        \protect\\
        $^\dagger$: Gu Wang and Fabian Manhardt have equally contributed. \\
        \corrAuthor: Xiangyang Ji is the corresponding author.
        \protect\\
        Recommended for acceptance by L. Liu, T. Hospedales, Y. LeCun, M. Long, J. Luo, W. Ouyang, M. Pietikinen, and T. Tuytelaars.
        \protect\\
        Digital Object Identifier no. 10.1109/TPAMI.2021.3136301
    }
}

% \markboth{IEEE TRANSACTIONS ON PATTERN ANALYSIS AND MACHINE INTELLIGENCE, VOL. X, NO. X, MMMMMMM YYYY}%
% {WANG \MakeLowercase{\textit{et al.}}: Occlusion-Aware Self-Supervised Monocular 6D Object Pose Estimation}
\markboth{IEEE TRANSACTIONS ON PATTERN ANALYSIS AND MACHINE INTELLIGENCE}%
{\MakeUppercase{WANG et al.: Occlusion-Aware Self-Supervised Monocular 6D Object Pose Estimation}}



\IEEEtitleabstractindextext{%
\input{sections/0_abstract}

}

\maketitle
\copyrightnotice

\IEEEdisplaynontitleabstractindextext

% \fi
\IEEEpeerreviewmaketitle

\input{sections/1_intro}
\input{sections/2_related}

\input{sections/3_method}

\input{sections/4_exp}
\input{sections/5_conclusion} %

% or


%\appendices

%\section{}

% \ifCLASSOPTIONcompsoc
%   % The Computer Society usually uses the plural form
%   \section*{Acknowledgments}
% \else
%   % regular IEEE prefers the singular form
% \fi

% Gu Wang and Fabian Manhardt have equally contributed.

\ifCLASSOPTIONcaptionsoff
  \newpage
\fi



%\bibliographystyle{IEEEtran}
%\begin{thebibliography}{1}


%\end{thebibliography}
% NOTE: if references are broken, change to XeLatex, compile, then change back to pdfLatex, compile
\bibliographystyle{IEEEtran}
\bibliography{tpami_conf_journal_names,self6dpp_refs}

\vfill



\end{document}

%% file: packages.tex
\ifCLASSOPTIONcompsoc
  \usepackage[nocompress]{cite}
\else
  \usepackage{cite}
\fi
\usepackage{color,xcolor}
\usepackage{epsfig}
\usepackage{graphicx}
\usepackage{textcomp}
\usepackage{gensymb}
\usepackage{mathtools}
\usepackage{comment}

\usepackage{colortbl}
\usepackage[flushleft]{threeparttable}
\usepackage[export]{adjustbox}
\usepackage{array}
\usepackage{booktabs}
\usepackage{wrapfig}
\usepackage{hhline}
\usepackage{multirow}
\usepackage{ragged2e}
\usepackage{tabularx}
\usepackage{tabulary}
\newcolumntype{x}[1]{>{\centering\arraybackslash}p{#1pt}}
\usepackage{amsmath,amsfonts,amsthm,amssymb}
\usepackage{bm}
\usepackage{nicefrac}
\usepackage{microtype}
\usepackage[misc,geometry]{ifsym} 
\usepackage{pifont}% http://ctan.org/pkg/pifont
\newcommand{\cmark}{\ding{51}}%
\newcommand{\xmark}{\ding{55}}%

\usepackage{changepage}
\usepackage{extramarks}
\usepackage{fancyhdr}
\usepackage{lastpage}
\usepackage{setspace}
\usepackage{soul}
\usepackage{xspace}

\usepackage{url}
\usepackage{footmisc}

\usepackage{algorithm, algorithmic}
\usepackage{enumerate}

\usepackage{bbm}

\usepackage{orcidlink}
\usepackage{scalerel}
\usepackage{tikz}
\usetikzlibrary{svg.path}

\definecolor{orcidlogocol}{HTML}{A6CE39}
\tikzset{
    orcidlogo/.pic={
        \fill[orcidlogocol] svg{M256,128c0,70.7-57.3,128-128,128C57.3,256,0,198.7,0,128C0,57.3,57.3,0,128,0C198.7,0,256,57.3,256,128z};
        \fill[white] svg{M86.3,186.2H70.9V79.1h15.4v48.4V186.2z}
        svg{M108.9,79.1h41.6c39.6,0,57,28.3,57,53.6c0,27.5-21.5,53.6-56.8,53.6h-41.8V79.1z M124.3,172.4h24.5c34.9,0,42.9-26.5,42.9-39.7c0-21.5-13.7-39.7-43.7-39.7h-23.7V172.4z}
        svg{M88.7,56.8c0,5.5-4.5,10.1-10.1,10.1c-5.6,0-10.1-4.6-10.1-10.1c0-5.6,4.5-10.1,10.1-10.1C84.2,46.7,88.7,51.3,88.7,56.8z};
    }
}

\newcommand\orcidicon[1]{\href{https://orcid.org/#1}{\mbox{\scalerel*{
                \begin{tikzpicture}[yscale=-1,transform shape]
                \pic{orcidlogo};
                \end{tikzpicture}
            }{|}}}}

%% file: macros.tex
\newcolumntype{L}[1]{>{\raggedright\let\newline\\\arraybackslash\hspace{0pt}}m{#1}}
\newcolumntype{C}[1]{>{\centering\let\newline\\\arraybackslash\hspace{0pt}}m{#1}}
\newcolumntype{R}[1]{>{\raggedleft\let\newline\\\arraybackslash\hspace{0pt}}m{#1}}

\newcommand{\app}{\raise.17ex\hbox{$\scriptstyle\sim$}}
\newcommand*{\x}{\mathsf{x}\mskip1mu}

\newlength\savewidth\newcommand\shline{\noalign{\global\savewidth\arrayrulewidth
  \global\arrayrulewidth 1pt}\hline\noalign{\global\arrayrulewidth\savewidth}}
\newcommand{\tablestyle}[2]{\setlength{\tabcolsep}{#1}\renewcommand{\arraystretch}{#2}\centering\footnotesize}

\newcommand{\tabincell}[2]{\begin{tabular}{@{}#1@{}}#2\end{tabular}}

\newcommand{\eqn}[1]{Eq.~(\ref{#1})}
\newcommand{\fig}[1]{Fig.~\ref{#1}}
\newcommand{\tbl}[1]{TABLE~\ref{#1}}

\newcommand{\ignorethis}[1]{}

\makeatletter
\DeclareRobustCommand\onedot{\futurelet\@let@token\@onedot}
\def\@onedot{\ifx\@let@token.\else.\null\fi\xspace}

\def\eg{\emph{e.g}\onedot} 
\def\ie{\emph{i.e}\onedot} 
\def\cf{\emph{c.f}\onedot} 
 \def\vs{\emph{vs}\onedot}
\def\wrt{w.r.t\onedot} 
 \def\etal{\emph{et al}\onedot}
\makeatother

\definecolor{MyDarkBlue}{rgb}{0,0.08,1}
\definecolor{MyDarkGreen}{rgb}{0.02,0.6,0.02}
\definecolor{MyDarkRed}{rgb}{0.8,0.02,0.02}
\definecolor{MyDarkOrange}{rgb}{0.40,0.2,0.02}
\definecolor{MyPurple}{RGB}{111,0,255}
\definecolor{MyRed}{rgb}{1.0,0.0,0.0}
\definecolor{MyGold}{rgb}{0.75,0.6,0.12}
\definecolor{MyDarkgray}{rgb}{0.66, 0.66, 0.66}

\newcommand{\ignore}[1]{}

\definecolor{RoneColorBG}{rgb}{1, 1, 1}
\definecolor{RtwoColorBG}{rgb}{1, 1, 1}
\definecolor{RthreeColorBG}{rgb}{1, 1, 1}
\definecolor{RoneColor}{rgb}{0, 0, 0}
\definecolor{RtwoColor}{rgb}{0, 0, 0}
\definecolor{RthreeColor}{rgb}{0, 0, 0}

\newcommand{\Rone}[1]{\textcolor{RoneColor}{#1}}
\newcommand{\Rtwo}[1]{\textcolor{RtwoColor}{#1}}
\newcommand{\Rthree}[1]{\textcolor{RthreeColor}{#1}}

\def\loss{\mathcal{L}}

\newcommand{\myparagraph}[1]{\vspace{5pt}\noindent\emph{#1}\hspace{10pt}}

\newcommand\Set[2]{\left\{\,#1\mid#2\,\right\}}

\DeclareMathOperator{\avg}{avg}

\def\pselfvo{\text{Self6D}}

\def\pself{\text{Self6D++}}

\def\ours{\text{OURS}}
\def\oursLB{\text{OURS\textsubscript{(LB)}}}
\def\oursUB{\text{OURS\textsuperscript{(UB)}}}

\def\Dref{D_\text{ref}}

\def\pseudoMfull{\widetilde{M}_\text{amodal}}
\def\pseudoMvis{\widetilde{M}_\text{vis}}
\def\pseudoM{\widetilde{M}}

\def\predMfull{\widehat{M}_\text{amodal}}
\def\predMvis{\widehat{M}_\text{vis}}

\def\Mfull{M_\text{amodal}}
\def\Mvis{M_\text{vis}}

\def\pseudoPose{\widetilde{P}}

\def\pseudoPoseInit{\widetilde{P}_\text{init}}
\def\pseudoRotInit{\widetilde{R}_\text{init}}
\def\pseudoTransInit{\widetilde{t}_\text{init}}

\def\predPose{\widehat{P}}
\def\predRot{\widehat{R}}
\def\predTrans{\widehat{t}}

\newcommand{\winNoPose}[1]{\textbf{#1}}
\newcommand{\winAll}[1]{\underline{#1}}
\newcommand{\winAllwoPose}[1]{\underline{\textbf{#1}}}

\newcommand{\corrAuthor}{$^{\textrm{\Letter}}$}

%% file: sections/0_abstract.tex
\begin{abstract} % ---------------------------------------------------------------
\justifying
6D object pose estimation is a fundamental yet challenging problem in computer vision. Convolutional Neural Networks (CNNs) have recently proven to be capable of predicting reliable 6D pose estimates even under monocular settings. Nonetheless, CNNs are identified as being extremely data-driven, and acquiring adequate annotations is oftentimes very time-consuming and labor intensive. To overcome this limitation, we propose a novel monocular 6D pose estimation approach by means of self-supervised learning, removing the need for real annotations. After training our proposed network fully supervised with synthetic RGB data, we leverage current trends in noisy student training and differentiable rendering to further self-supervise the model on these unsupervised real RGB(-D) samples, seeking for a visually and geometrically optimal alignment. Moreover, employing both visible and amodal mask information, our self-supervision becomes very robust towards challenging scenarios such as occlusion. Extensive evaluations demonstrate that our proposed self-supervision outperforms all other methods relying on synthetic data or employing elaborate techniques from the domain adaptation realm. Noteworthy, our self-supervised approach consistently improves over its synthetically trained baseline and often almost closes the gap towards its fully supervised counterpart. The code and models are publicly available at \url{https://github.com/THU-DA-6D-Pose-Group/self6dpp.git}.
% \href{https://github.com/THU-DA-6D-Pose-Group/self6dpp.git}{https://github.com/THU-DA-6D-Pose-Group/self6dpp.git}.
\end{abstract}

%% file: sections/1_intro.tex
\section{Introduction}
\IEEEPARstart{M}{onocular} estimation of the 6D pose (\ie~3D translation and 3D rotation) of objects \wrt the camera is a long-standing problem in computer vision.
Accurately localizing the 6D object pose is crucial for a wide range of real-world applications such as robotic manipulation~\cite{collet2011moped,Zhu2014SingleI3,tremblay2018deep},
augmented reality~\cite{marchand2015pose,su2019deep}, and autonomous driving~\cite{geiger2012we,manhardt2019roi}.
Learning-based works have recently shown very promising results for the task at hand.
Nonetheless, these methods are typically fed with a huge amount of data during training~\cite{Wang_2021_GDRN,hodan2018bop}.
Yet, acquiring appropriate training labels is very time-consuming and labor intensive~\cite{kaskman2019homebreweddb,hodan2019photorealistic}.
This is particularly true for 6D pose estimation, as, for each RGB-D frame, 3D CAD models need to be precisely aligned across a difficult and tedious process.

As a consequence, several different approaches to tackle the lacking of real labels have been proposed in literature.
The most common way is to simply render a huge number of synthetic training images with tools such as OpenGL~\cite{richter2016playing,Su_2015_ICCV} \Rthree{or Blender~\cite{peng2019pvnet}}.
Via sampling of random 6D poses, large amounts of synthetic images can be simulated using the accompanying 3D CAD models.
In addition, it is common to employ domain randomization to impose invariance to changing scenes~\cite{kehl2017ssd,manhardt2018deep}. Thereby, images from large-scale 2D object datasets such as VOC~\cite{everingham2010pascal} and COCO~\cite{coco_eccv14} are utilized as background
for the rendered samples.
Another line of work attempts to conduct domain adaptation by means of techniques such as GANs to translate the input from source to target domain or vice-versa~\cite{bousmalis2017unsupervisedPixelda}.
Nonetheless, as these rendered images can be still effortlessly distinguished from real imagery due to low quality and physical implausibility,
recent works instead investigated the use of physically-based rendering to increase quality and additionally enforce real physical constraints~\cite{hodan2019photorealistic,tremblay2018deep}.
Eventually, despite techniques based on domain adaption~\cite{bousmalis2017unsupervisedPixelda}, domain randomization~\cite{sundermeyer2018implicit} and photorealistic rendering~\cite{hodan2019photorealistic} are steadily reducing the synthetic-to-real domain gap,
the performance is still far from comparable with respect to methods exploiting real imagery with 6D pose annotations during training.

\input{figs/0_intro_fig}

Inspired by current trends from differentiable rendering~\cite{liu2019softras,kato2018renderer} and self-supervised learning~\cite{godard2017unsupervised,cmrKanazawa18},
we want to tackle the problem of lack of real labeled data from an entirely different viewpoint.
Humans have the amazing ability to learn how to reason about the 3D world from 2D images alone.
Moreover, we can even learn 3D world properties without requiring another human or \textit{labels} in a self-supervised fashion
by validating if observations of the world agree with the anticipated outcome.
For instance, infants learn about \emph{what is a 3D object} simply by observing which things move together~\cite{spelke1990principles}.
Translating this to the task of 6D pose estimation -- while labeling the pose is a clear obstacle,
obtaining real-world observations in form of RGB(-D) images can be easily accomplished at scale.
Hence, as humans, we want the network to be capable of learning by itself from these unlabeled examples in a self-supervised fashion.

To establish supervision, the method is required to understand the 3D world solely from 2/2.5D data.
Experiencing the 3D world in form of 2D images on the eye's retina is known as \textit{rendering} and
is a very active field of research in Computer Graphics~\cite{marschner2015CG}.
Unfortunately, most standard rendering pipelines for 3D CAD models are not differentiable as they rely on rasterization and
gradients cannot be calculated for the $argmax$ function.
To circumvent this issue, several ideas have been recently proposed to re-establish the gradient flow for rendering~\cite{kato2020differentiable}.
Thereby, \Rtwo{most methods either approximate} a \emph{``useful''} gradient~\cite{kato2018renderer,opendr_eccv14},
or compute the analytical gradient by approximating the rasterization function itself~\cite{liu2019softras,chen2019learning_dibrenderer}.
Nevertheless, to obtain such useful gradient, the rendered and real images should be fairly similar.
To this end, we adopt a two-stage method as illustrated in Fig.~\ref{fig:intro}.
In particular, we first train our model fully-supervised on physical-based renderings~\cite{hodan_bop20}.
We then employ unannotated RGB(-D) data to self-supervise the model on unseen real data using differentiable rendering,
in an effort to diminish the synthetic-to-real domain gap.

To summarize, we make the following main technical contributions.
\begin{itemize}
\item We propose a self-supervised method for monocular 6D object pose estimation that leverages unsupervised RGB (or additionally with depth) images to domain adapt the model based on a differentiable learning pipeline.
\item We leverage both visible and amodal object mask prediction to develop an occlusion-aware pose estimator and establish different self-supervised loss terms via visual and geometric alignment.
\item We further exploit noisy student training~\cite{Xie_2020_CVPR_noisy_student} and an RGB-based deep pose refiner~\cite{li2019deepim,labbe2020cosypose} to improve the overall pose accuracy as well as its robustness to different nuisances such as occlusion and photometric changes.
\end{itemize}

To the best of our knowledge, we are the first to conduct self-supervised 6D object pose estimation from real data, without the need of real-world 6D labels.
We experimentally demonstrate that the proposed self-supervised approach,
which we dub $\pself$, outperforms state-of-the-art methods for monocular 6D object pose estimation trained without real annotations by a large margin.

A preliminary version of this work ($\pselfvo$) has been published as oral in ECCV 2020~\cite{wang2020self6d}.
In this work, we present our revised $\pself$ featuring several improvements compared to the original version.
First, we replace the single-stage pose estimator by a stronger two-stage approach, notably improving over our baseline.
As Yolov4~\cite{bochkovskiy2020yolov4} provides more robust 2D detections, we can also obtain much more accurate poses using our improved version of the state-of-the-art 6D pose regressor GDR-Net~\cite{Wang_2021_GDRN}.
Second, we introduce several important modifications to improve self-supervision under severe occlusion.
In particular, the pose predictor is extended to predict both visible and full amodal masks for self-supervision.
We also leverage noisy student training~\cite{Xie_2020_CVPR_noisy_student} in the self-supervised training phase to enhance robustness.
We demonstrate that, differently from~\cite{wang2020self6d}, these contributions are able to release the constraint of having depth data available during self supervision, making our method fully monocular.
To this end, we employ the RGB-based deep pose refiner from~\cite{li2019deepim,labbe2020cosypose} for the teacher model so to obtain more reliable pseudo pose labels.
Finally, we demonstrate the superiority of our approach through extensive experiments.
Particularly, we have not only significantly improved our baseline in \cite{wang2020self6d} on common benchmark datasets including LINEMOD~\cite{Hinterstoisser2012}, HomebrewedDB~\cite{kaskman2019homebreweddb}, Cropped LINEMOD~\cite{bousmalis2017unsupervisedPixelda} and Occluded LINEMOD~\cite{Brachmann2014Learning6O}, but also added the new evaluation for YCB-Video~\cite{xiang2017posecnn}.
Moreover, compared to the initial version~\cite{wang2020self6d}, our $\pself$ is able to produce more reliable poses under challenging scenarios \wrt Occluded LINEMOD and YCB-Video, which exhibit many symmetric objects undergoing significant occlusion.
Throughout all benchmarks, we can almost completely fill up the gap with state-of-the-art fully-supervised methods employing real-world 6D pose labels.

%% file: figs/0_intro_fig.tex
\begin{figure}[t!]
	\centering
	\includegraphics[width = 0.99\linewidth]{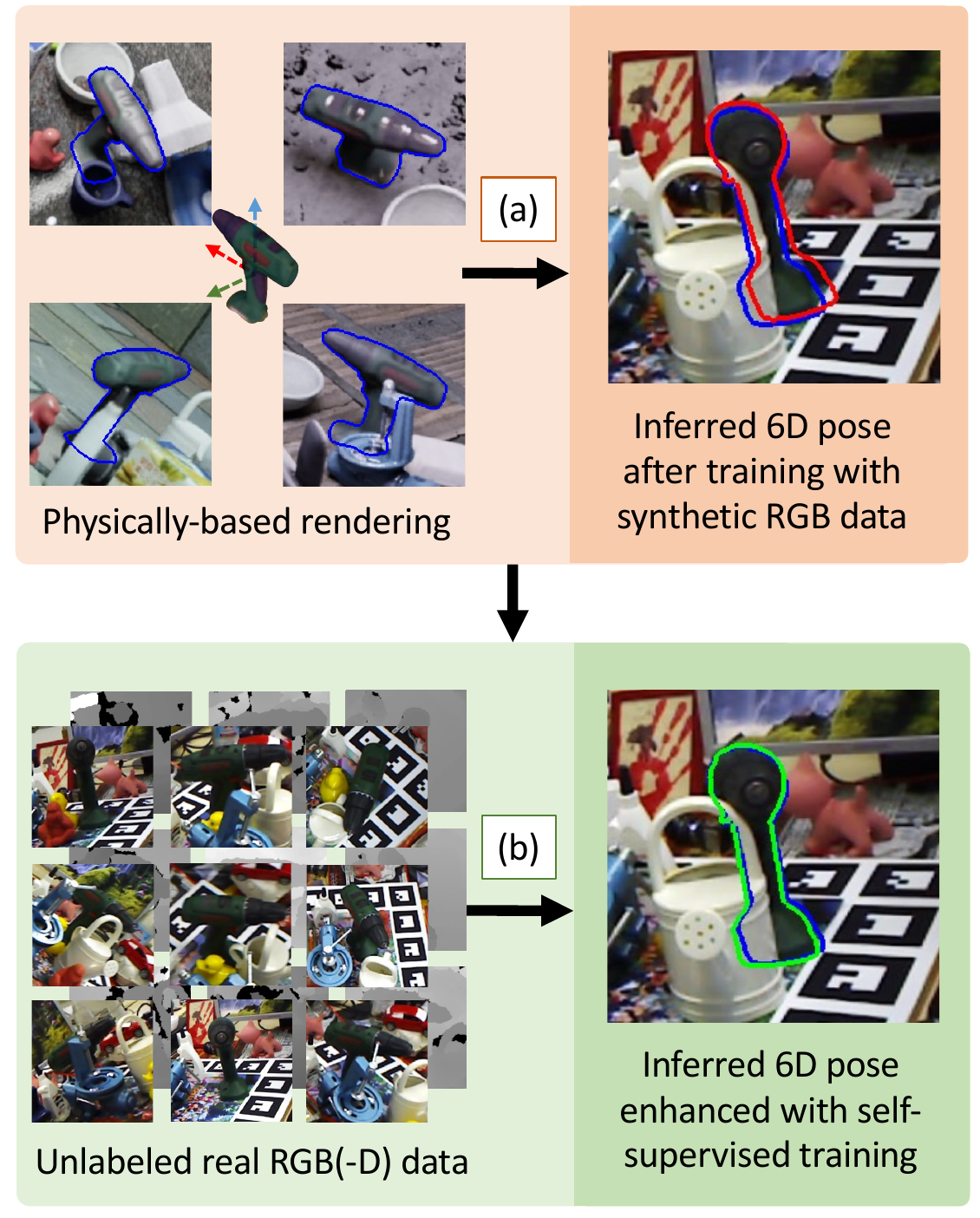}
	\caption{\emph{Abstract illustration} of our proposed methodology.
	To circumvent the use of real 6D pose annotations, we initially train our model purely on synthetic RGB data (\textit{a}).
	Secondly, employing a large amount of unlabeled real RGB (and \textit{optionally depth}) images (\textit{b}), we significantly improve performance (\textit{right}).
	The \textit{blue}, \textit{red} and \textit{green} silhouettes respectively represent the ground-truth 6D pose, the results before and after applying our self-supervision.
	}
	\label{fig:intro}
\end{figure}

%% file: sections/2_related.tex
\section{Related work}
We first introduce recent works in monocular 6D pose estimation.
Afterwards, we discuss important methods from differentiable rendering as they form the core of our (as well as other) self-supervised learning frameworks.
We then outline other successful approaches grounded on self-supervised learning.
Lastly, we take a brief look at domain adaptation in the field of 6D pose, since our method can be considered as an implicit formulation to close the synthetic-to-real domain gap.

\subsection{Monocular 6D Pose Estimation}
Traditional object pose estimation approaches rely on local image features~\cite{lowe1999object,Romea-2011-7355} or template matching~\cite{hinterstoisser2012gradient}, assuming a grayscale or RGB images as input.
With the introduction of consumer RGB-D cameras, attention shifted more towards conducting object pose estimation from depth or RGB-D images.
It is true that methods leveraging RGB-D data within template matching~\cite{Hinterstoisser2012},  point pair features~\cite{hinterstoisser2016going} and learning-based methods ~\cite{brachmann2014learning,krull2015learning} often yield superior performance over RGB-only counterparts, however, they also restricted in applications as depth sensors have apparent limitations including frame rate, field of view, resolution, depth range, power consumption, and noise due to \eg reflections or scattering of the laser.

Henceforth, CNN-based monocular 6D pose estimation has received a lot of attention and several very promising works have been proposed~\cite{hodan2018bop}. In general there are three different branches that have been extensively explored in recent history.

One major branch is grounded on establishing 2D-3D correspondences between the image and the 3D CAD model.
After estimating these correspondences, P$n$P is commonly employed to solve for the 6D pose.
For instance, \Rthree{BB8~\cite{rad2017bb8} and YOLO6D~\cite{tekin18_yolo6d}} propose to employ a CNN to estimate the 2D projections of the 3D bounding box corners in image space.
Similarly, \Rthree{SegDriven~\cite{hu2019segpose} and PVNet~\cite{peng2019pvnet}} also regress 2D projections of associated sparse 3D keypoints,
however, both employ segmentation paired with voting to improve reliability.
\Rthree{HybridPose~\cite{song2020hybridpose}} harnesses several intermediate representations for the 6D pose.
In particular, Song \etal infer sparse 2D-3D correspondences together with edge vectors and symmetry correspondences to increase robustness by means of a more diverse set of features.
In contrast, \Rthree{DPOD~\cite{zakharov2019dpod}, CDPN~\cite{li2019cdpn}, Pix2Pose~\cite{park2019pix2pose}, and EPOS~\cite{hodan2020epos}} ascertain dense 2D-3D correspondences, rather than sparse ones.
While \Rthree{CDPN~\cite{li2019cdpn}} decouples the estimation of translation and rotation, \Rthree{Pix2Pose~\cite{park2019pix2pose}} leverages adversarial training~\cite{NIPS2014_GAN} to better cope with occlusion.
\Rthree{EPOS~\cite{hodan2020epos}} establishes dense correspondences in an ambiguity-free manner.
Using classification of fragments rather than directly regressing the correspondences, ambiguities can be deduced from the predicted distributions during inference.

Another branch of works directly regress the 6D pose.
For instance, while \Rthree{SSD6D~\cite{kehl2017ssd} extends SSD~\cite{liu2016ssd}} to also classify the viewpoint and in-plane rotation,
\Rthree{MHP~\cite{manhardt2019ambiguity}} further adjusts \Rthree{SSD6D~\cite{kehl2017ssd}} to implicitly deal with ambiguities by means of multiple hypotheses.
In \Rthree{PoseCNN~\cite{xiang2017posecnn} and DeepIM~\cite{li2019deepim}} the authors minimize a point matching loss.
\Rthree{CosyPose~\cite{labbe2020cosypose} extends DeepIM~\cite{li2019deepim}} by the help of two cascaded iterative pose refiners for coarse registration and coarse-to-fine alignment, respectively.
In addition, pose graph consistency of different static objects is enforced from multiple views to obtain globally optimal solutions.
Although inferring 2D-3D correspondences, \Rthree{SingleStage~\cite{hu2020single} and GDR-Net~\cite{Wang_2021_GDRN}} directly estimate the 6D pose via learning of the P$n$P paradigm.
Both show that learned P$n$P can produce more robust estimates than standard P$n$P, especially when the objects of interest are exposed to occlusions.

A handful of methods instead learn a pose embedding, which can be subsequently utilized for retrieval of pose.
In particular, inspired by~\cite{wohlhart2015learning,kehl2016deep},
Sundermeyer \etal~\cite{sundermeyer2018implicit} employ an Augmented AutoEncoder (AAE) to learn latent representations for the 3D rotation.
\Rthree{PAE~\cite{li2020PAE}} further exploits dense coordinates rather than RGB values as the reconstruction is thus enforced to better account for pose ambiguity.
Whereas these works train separate encoders for individual objects, \Rthree{Sundermeyer \etal~\cite{Sundermeyer_2020_MultiPath}} propose to learn a shared encoder together with multiple decoders to efficiently build latent embeddings for various different objects.

Noteworthy, the majority of these methods~\cite{hu2019segpose,park2019pix2pose,rad2017bb8,tekin18_yolo6d,xiang2017posecnn} exploit annotated real data to train their models.
However, labeling real data commonly comes with huge efforts in time and labor.
Moreover, a shortage of sufficient real-world annotations can lead to overfitting, regardless of exploiting strategies such as Cut\&Paste~\cite{dwibedi2017cut,kaskman2019homebreweddb}.
Other works, in contrast, fully rely on synthetic data to deal with these pitfalls~\cite{sundermeyer2018implicit,manhardt2019ambiguity}.
Nonetheless, the performance falls far behind methods using real pose labels.
We, hence, harness the best of both worlds.
While unannotated data can be easily obtained at scale, this in combination with our self-supervision for pose is able to outperform all methods trained on synthetic data by a large margin.

% ----------------------------------------------
\subsection{Differentiable Rendering}
Optimizing parameters through rendering has been proposed for several representations including voxel maps~\cite{yan2016perspective,tulsiani2017multi}, pointclouds~\cite{wiles2020synsin,insafutdinov2018unsupervised}, implicit functions~\cite{Zakharov_2020_CVPR,mildenhall2020nerf}, and 3D meshes~\cite{kato2018renderer,chen2019learning_dibrenderer}.
In 6D pose estimation, one usually infers the orientation and translation given a 3D CAD model,
therefore, in this section we focus on non-parametric differentiable rendering methods for 3D meshes instead of parametric methods which attempt to learn the rendering function through neural networks~\cite{tewari2020state}.
There are two separate lines of work for differentiably rendering of 3D meshes, \ie, rasterization~\cite{opendr_eccv14,kato2018renderer} as well as ray-tracing~\cite{li2018differentiable}.
Since the latter is computationally much more expensive, we concentrate on rasterization-driven approaches here.

Rasterization involves discrete assignment operations, preventing the flow of gradients through the rendering process.
A series of work have been devoted to circumvent the hard assignment in order to reestablish the gradient flow.
\Rthree{Loper \etal~\cite{opendr_eccv14}} introduce the first differentiable renderer, namely OpenDR,
by means of calculating the derivative of pixel values \wrt the 2D pixel positions within the image-space via first-order Taylor approximation.
However, in OpenDR a vertex can only receive gradients from neighboring pixels within a close range of the mesh face.
In \Rthree{NMD~\cite{kato2018renderer}}, the authors instead approximate the gradient as the potential change of the pixel's intensity \wrt the meshes' vertices.
SoftRas~\cite{liu2019softras} conducts rendering by aggregating the probabilistic contributions of each mesh triangle in relation to the rendered pixels.
Consequently, the gradients can be calculated analytically, however, with the cost of extra computation and a loss in image quality.
\emph{DIB-R}~\cite{chen2019learning_dibrenderer} further extends \Rthree{SoftRas~\cite{liu2019softras}} by considering rasterization as
a combination of weighted interpolation of local mesh properties for foreground and global aggregation for background,
which yields clearer images whilst still allowing occluded vertices to contribute to the optimization.
In this work, we use \emph{DIB-R}~\cite{chen2019learning_dibrenderer} since it can be considered state-of-the-art for differentiable rendering.
Moreover, we extend \emph{DIB-R} such that it also renders the accompanying depth map~\cite{wang2020self6d}.

%-----------------------------------------------
\subsection{Recent Trends in Self-Supervised Learning}
Self-supervised learning, \ie~learning despite the lack of properly labeled data,
has recently enabled a large number of applications ranging from 2D image understanding all the way down to depth estimation for autonomous driving.
In the core, self-supervised learning approaches implicitly learn about a specific task through solving related proxy tasks.
This is commonly achieved by enforcing different constraints such as pixel consistencies across multiple views or modalities.

One prominent approach in this area is MonoDepth~\cite{godard2017unsupervised},
which conducts monocular depth estimation
by warping the 2D image points into another view, enforcing a minimum reprojection loss.
In the following many works to extend MonoDepth have been introduced~\cite{pillai2019superdepth,godard2019digging,guizilini20203d}.
In visual representation learning, consistency is ensured by solving pretext tasks~\cite{kolesnikov2019revisiting} or contrastive learning~\cite{He_2020_CVPR_Moco,chen2021CVPR_SimSiam}.
Another line of works explore self-supervised learning for 3D human pose estimation,
leveraging multi-view epipolar geometry~\cite{kocabas2019self} or imposing 2D-3D consistency via lifting and reprojection of keypoints~\cite{chen2019unsupervised}.
Self-supervised learning approaches using differentiable rendering have also been proposed in the field of 3D object and human body reconstruction from single RGB images~\cite{tung2017self,cmrKanazawa18,omran2018neural,alldieck2019learning,Zuffi_2019_ICCV}.

In the domain of 6D pose estimation, self-supervised learning is still a rather less explored field.
Deng \etal~\cite{deng2020self} propose a novel self-labeling pipeline with an interactive robotic manipulator.
Essentially, running several methods for 6D pose estimation, they can reliably generate precise annotations.
Nonetheless, the final 6D pose estimation model is still trained fully-supervised using the acquired data.
\Rthree{Zakharov \etal~\cite{Zakharov_2020_CVPR}} employ differentiable rendering of implicit functions to auto-label 3D bounding boxes on KITTI3D~\cite{geiger2012we}.
Nevertheless, similar to \Rthree{Deng \etal~\cite{deng2020self}}, the final pose estimator is obtained via fully supervised training on the produced labels.

The preliminary version of our work~\cite{wang2020self6d} is the first method which proposes to instead directly establish self-supervision for 6D pose by enforcing visual and geometric consistencies on top of unlabeled RGB-D images on the ground of differentiable rendering.
In the meanwhile, \Rthree{Beker \etal~\cite{beker2020self}} harness multiple cues such as object detection, object segmentation, and depth prediction to self-supervise the model with differentiable rendering.
However, a separate model needs to be optimized for each detection,
which makes the approach slow due to the computational overhead for rendering and gradient computation.
More recently, Sock \etal~\cite{sock2020} also establish self-supervision with unlabeled images using differentiable rendering,
however, only rely on RGB data.
In addition, inspired by works on self-supervised depth estimation, multi-view consistencies are employed to improve pose quality.

\input{figs/1_arch}
% ------------------------------------
\subsection{Domain Adaptation for 6D Pose Estimation}
Bridging the domain gap between synthetic and real data is crucial in 6D pose estimation.
Many works tackle this problem
by learning a transformation to align the synthetic and real domains
via Generative Adversarial Networks (GANs) ~\cite{bousmalis2017unsupervisedPixelda,lee2018diverse,zakharov2019deceptionnet}
or by means of feature mapping~\cite{rad2018domain}.
Exemplary, \Rthree{Lee \etal~\cite{lee2018diverse}} use a cross-cycle consistency loss based on disentangled representations to embed images onto a domain-invariant content space and a domain-specific attribute space.
Rad \etal~\cite{rad2018domain} instead translate features from a color-based pose estimator to a depth-based pose estimator.

In contrast, works from domain randomization aim at learning domain-invariant attributes.
This can be accomplished harnessing random backgrounds and severe augmentations~\cite{kehl2017ssd,sundermeyer2018implicit} or
employing CNNs to generate novel backgrounds and image augmentations~\cite{zakharov2019deceptionnet}.
While \Rthree{SSD6D~\cite{kehl2017ssd} and AAE~\cite{sundermeyer2018implicit}} harness COCO images as background together with various augmentations to become invariant to the domain,
\Rthree{DeceptionNet~\cite{zakharov2019deceptionnet}} employs adversarial training to generate backgrounds and image augmentations, maximally fooling the pose estimation network.

%% file: figs/1_arch.tex
\begin{figure*}[t!]
	\centering
	\includegraphics[width = 0.99\linewidth]{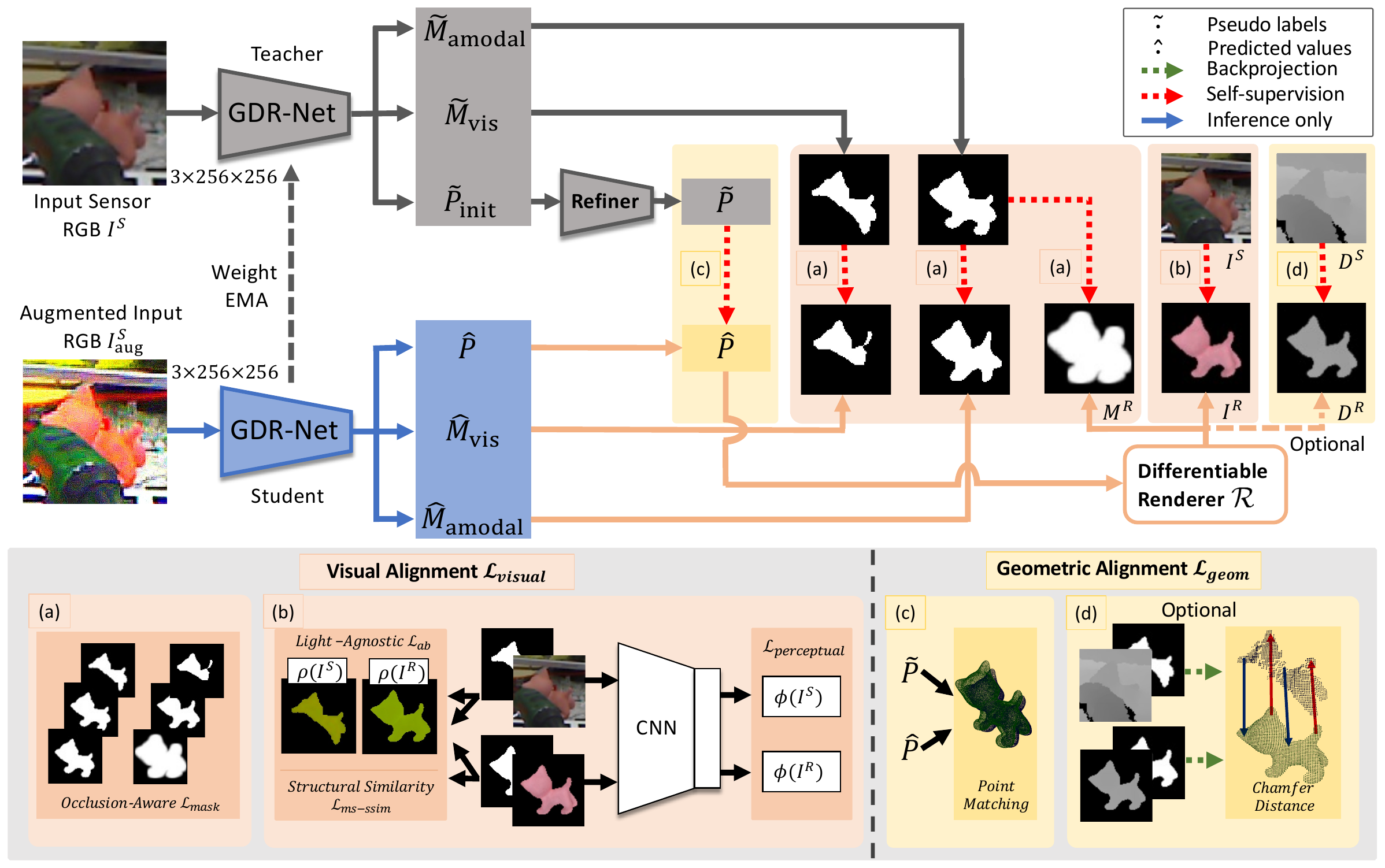}
	\caption{\Rone{\emph{Self-supervising monocular 6D object pose estimation.}
	 \textit{Top}: After training of the Yolov4~\cite{bochkovskiy2020yolov4} object detector (omitted for clarity), the extended GDR-Net~\cite{Wang_2021_GDRN} pose estimator, and the pose refiner purely with synthetic RGB images, we leverage noisy student training and differentiable rendering to self-supervise the pose estimator using a large amount of unlabeled RGB(-D) images $(I^S, D^S)$. 
	 The predicted values $(\predPose, \predMvis, \predMfull)$ of GDR-Net from the augmented RGB input $I_{\text{aug}}^S$ are self-supervised by the pseudo labels $(\pseudoPose, \pseudoMvis, \pseudoMfull)$ from the clean input RGB $I^S$.
	 We also differentiably render ($\mathcal{R}$) the associated RGB(-D) image and \Rthree{probabilistic} mask $(I^R, D^R, M^R)$ to register against the sensor RGB(-D) data and the pseudo amodal mask $(I^S, D^S, \pseudoMfull)$.
	\textit{Bottom}: We impose various constraints to visually (\emph{a} and \emph{b}), and geometrically (\emph{c} and \emph{d}) optimize the 6D object pose without requiring real pose labels.}}
	\label{fig:arch}
\end{figure*}

%% file: sections/3_method.tex
\section{Methodology}
In this paper, we aim at conducting 6D pose estimation from monocular images without the need for a largely annotated dataset of real samples. 
To this end, we propose a novel method that can learn monocular pose estimation from both synthetic RGB data and unlabeled real-world RGB(-D) samples.
As illustrated in Fig.~\ref{fig:intro}, our approach is composed of two stages. 
Since generation of large-scale synthetic data can be obtained very easily at scale, we first train our model fully-supervised using synthetic RGB data only. 
Subsequently, to circumvent the issue of overfitting to the synthetic domain, we then enhance generalizability via self-supervision on unlabeled real-world RGB(-D) data. 

Fig.~\ref{fig:arch} summarizes the proposed approach for self-supervision of monocular 6D pose estimation \wrt unlabeled data.
We tackle the problem by means of establishing different visual and geometric constraints seeking the best alignment in terms of 6D pose. 
Employing the noisy student training paradigm and differentiable rendering, various visual and geometric consistencies can be established which in turn serve as strong error signal.

Contrary to \cite{wang2020self6d}, we make use of an occlusion-aware pose estimator built on top of GDR-Net~\cite{Wang_2021_GDRN}. 
We additionally harness visible and amodal object masks ($M_\text{vis}, M_\text{amodal}$) during self-supervision, to further improve robustness towards occlusion.
\Rtwo{Given a raw sensor RGB image} $I^S$ (and optionally the associated sensor depth image $D^S$), we first extract the object of interest using an off-the-shelf object detector such as Yolov4~\cite{bochkovskiy2020yolov4} (omitted in Fig.~\ref{fig:arch} for clarity). 
Thereby, the object detector and the pose estimator are both pre-trained on synthetic RGB data.
From this we then initialize the teacher as well as the student pose estimators with the same weights as obtained after pre-training. 
While both are fed with the same input RGB patch $I^S$, the student's patch further undergoes various augmentations $I_\text{aug}^S$.
Given the corresponding patch, the teacher and student then respectively \Rtwo{predict the visible masks $\pseudoMvis$ and $\predMvis$}, amodal masks $\pseudoMfull$ and $\predMfull$, and associated poses $\pseudoPoseInit$ and $\predPose$.
Through the supervision from the teacher, the student network thus has to become agnostic to any variance as induced via augmentation.
To also directly learn from the raw data, we additionally run differentiable rendering to obtain the RGB(-D) image ($I^R, D^R$) and \Rthree{probabilistic} amodal mask $M^R$ associated to the student's predicted pose $\predPose =[\predRot | \predTrans]$, composed as the predicted rotation $\predRot$ and translation $\predTrans$,
using the extended \emph{DIB-R}\footnote{We extended \emph{DIB-R} to conduct real perspective projection and also provide the depth map fully differentiably. The code has been made publicly available at \href{https://git.io/Self6D-Diff-Renderer}{https://git.io/Self6D-Diff-Renderer}.}~\cite{chen2019learning_dibrenderer} renderer from our original work~\cite{wang2020self6d} with
\begin{equation}
 \mathcal{R}(\predPose, K, \mathcal{M}) = (I^R, D^R, M^R).
\end{equation}
Thereby, $\mathcal{M}$ denotes the given 3D CAD model and $K$ is the known camera intrinsic matrix.
Moreover, leveraging a RGB-based deep pose refiner $\Dref$~\cite{li2019deepim,labbe2020cosypose}, 
we further refine the obtained pseudo pose labels $\pseudoPose = \Dref(\pseudoPoseInit)$. 
Despite $\Dref$ being only trained on synthetic data, the iterative process is capable of further enhancing the quality of the pseudo label, even removing the need for depth data, which is a must in \cite{wang2020self6d}.
Finally, the visual and geometric alignment are established by using the pseudo labels ($\pseudoPose, \pseudoMvis, \pseudoMfull$) and the sensor data ($I^S, D^S$) to directly self-supervise
the predicted values ($\predPose, \pseudoMvis, \predMfull$) and through their corresponding differentiably rendered data ($I^R, D^R, M^R$).

% --------------------------------------------------------------------------------------------
\subsection{Monocular Pose Estimation Under Occlusion}
% --------------------------------------------------------------------------------------------
As we depend on differentiable rendering, inference of the 6D pose $\predPose$ has to be fully differentiable in order to allow backpropagation. 
While it is possible to obtain gradients for P$n$P~\cite{chen2020end} as well as RANSAC~\cite{brachmann2017dsac}, they also come with the burden of a high memory footprint and computational effort, rendering them impractical for our online learning formulation.
Thus, we cannot resort to methods based on establishing 2D-3D correspondences~\cite{peng2019pvnet,li2019cdpn,hodan2020epos}, despite those currently dominating the field.
In the first version of this work~\cite{wang2020self6d}, we proposed a single-stage pose estimator inspired by ROI-10D~\cite{manhardt2019roi} in order to directly estimate the rotation and translation parameters. 
However, especially when confronted with occlusion, the performance is far inferior to recent methods based on 2D-3D correspondences.
To this end, we employ a more recent two-stage regression-based approach GDR-Net~\cite{Wang_2021_GDRN}, which combines the two domains of correspondence-driven and direct regression-based methods, by utilizing dense correspondences as geometrical guidance during pose inference.

Nonetheless, the vanilla GDR-Net is still not perfectly suitable for our self-supervision, especially when the object undergoes severe occlusion.
Since the rendered mask $M^R$ is always un-occluded, the pseudo mask needs to be un-occluded as well. 
However, as GDR-Net only produces one output channel for the visible object mask $\Mvis$, we append a second mask prediction branch for the un-occluded amodal object mask $\Mfull$.
Notice that ground truth for both masks can be easily obtained from the simulator.
This allows the pose estimator to better account for occlusion during self-supervision.
Furthermore, we apply two additional minor changes to improve over~\cite{Wang_2021_GDRN}. 
First, we replace Faster R-CNN~\cite{ren2015faster} with the faster and stronger Yolov4 detector~\cite{bochkovskiy2020yolov4}. 
Second, we exchange ResNet-34~\cite{he2016deep} with a more recent ResNeSt-50~\cite{zhang2020resnest} backbone.

As aforementioned, we need good initial estimates to enable self-supervision. 
Hence, we pre-train both the object detector and the occlusion-aware pose estimator fully-supervised on simulated RGB images.
Thereby, in contrast to \cite{Wang_2021_GDRN}, we employ the binary cross entropy loss for mask prediction. 
For all other losses for 6D pose and geometric features, we utilize the same objective functions as proposed in the original GDR-Net. 
We kindly refer the readers to~\cite{Wang_2021_GDRN} for more details.

% --------------------------------------------------------------------------------------------
\subsection{Self-supervising Monocular 6D Pose Estimation Under Occlusion}
% --------------------------------------------------------------------------------------------
In this section, we describe the details of visual and geometric alignment by means of noisy student training and differentiable rendering for our occlusion-aware self-supervised training.
For simplicity of the following, we define all foreground and background pixels for a given mask $M$ as 
\begin{equation}
    Pos(M) = \Set{(i,j)}{\forall M(i,j)=1}
\end{equation}
and
\begin{equation}
Neg(M) = \Set{(i,j)}{\forall M(i,j)=0}.
\end{equation}

\myparagraph{\textbf{Visual Alignment for Self-Supervision.}}
The most intuitive way is to simply align the rendered image $I^R$ with the sensor image $I^S$, deploying directly a loss on both samples. 
However, as the domain gap between $I^S$ and $I^R$ turns out to be very large, this does not work well in practice. 
In particular, lighting changes as well as reflection and bad reconstruction quality (especially in terms of color) oftentimes cause a high error despite having good pose estimates, eventually leading to divergence in the optimization. 
Hence, in an effort to keep the domain gap as small as possible, we impose multiple constraints measuring different domain-independent properties. 
In particular, we assess different visual similarities \wrt mask, color, image structure, and high-level content.

Since object masks are naturally domain agnostic, they can provide a particularly strong supervision. 
As our data is unannotated we refer to our pseudo masks $(\pseudoMvis, \pseudoMfull)$ for weak supervision. 
However, due to imperfect predicted masks, we utilize a reweighted cross-entropy loss~\cite{jiang2019integral}, which recalibrates the weights of positive and negative regions
\begin{equation}\label{eq:loss_mask}
\begin{aligned}
\loss_{rwce}(\widetilde{M}, M) \coloneqq & - \frac{1}{|Pos(\widetilde{M})|} \sum_{j\in Pos(\pseudoM)} \pseudoM_j \log M_j \\
& - \frac{1}{|Neg(\pseudoM)|} \sum_{j\in Neg(\pseudoM)} \log (1 - M_j),
\end{aligned}
\end{equation}
where $\pseudoM$ is the pseudo mask and $M$ is the corresponding predicted or rendered mask. 
Specifically, we harness both visible and full object masks in order to account for occlusion in visual alignment
\begin{equation}\label{eq:loss_mask_all}
\begin{aligned}
\loss_{mask} \coloneqq 
&~   \lambda_1 \loss_{rwce}(\pseudoMfull, M^R) \\ 
&~ + \lambda_2 \loss_{rwce}(\pseudoMfull, \predMfull)  \\
&~ + \lambda_3 \loss_{rwce}(\pseudoMvis, \predMvis),
\end{aligned}
\end{equation}
which is balanced by $\lambda_1, \lambda_2$ and $\lambda_3$. Noticeably, $\pselfvo$ only enforces a loss on the visible masks with $\loss_{rwce}(\pseudoMvis, M^R)$, turning out less robust towards occlusion compared to our formulation based on amodal masks and noisy-student training.

Although masks are not suffering from the domain gap, they discard a lot of valuable information. 
In particular, color information is often the only guidance to disambiguate the 6D pose, especially for geometrically simple objects.
We thus resort to several techniques to mitigate the domain gap between rendered and real images.

Since the domain shift is at least partially caused by light, we attempt to decouple light prior to measuring color similarity. 
Let $\rho$ denote the transformation from RGB to LAB space, additionally discarding the light channel, we evaluate color coherence on the remaining two channels according to
\begin{equation}\label{eq:loss_AB}
\loss_{ab} \coloneqq 
\frac{1}{|Pos(\pseudoMvis)|} 
    \sum_{j \in Pos(\pseudoMvis)} 
    \|\rho(I^S)_j \cdot \pseudoM_{\text{vis}_j} - \rho(I^R)_j\|_1.
\end{equation}
 
We also avail various ideas from image reconstruction and domain translation, as they succumb the same dilemma.  
We assess the structural similarity (SSIM) in the RGB space and additionally follow the common practice to use a multi-scale variant, namely MS-SSIM~\cite{zhao2016loss}
\begin{equation}
\loss_{ms\text{-}ssim} \coloneqq 
     1 - ms\text{-}ssim(I^S \odot \pseudoMvis, I^R, s).
\end{equation}
Thereby, $\odot$ denotes the element-wise multiplication and $s=5$ is the number of employed scales. 
For more details on MS-SSIM, we kindly refer the readers to ~\cite{zhao2016loss}.

Another common practice is to appraise the perceptual similarity~\cite{johnson2016perceptual,zhang2018unreasonable} in feature space. 
To this end, a pre-trained deep neural network such as AlexNet~\cite{krizhevsky2012imagenet} is commonly employed to ensure low- and high-level similarity. 
We apply the perceptual loss at different levels of AlexNet. 
Specifically, we extract the feature maps of $L=5$ layers and normalize them along the channel dimension. 
Then we compute squared $L_2$ distances of the normalized feature maps $\phi^{(l)}(\cdot)$ for each layer $l$.
We average the individual contributions spatially and sum across all layers~\cite{zhang2018unreasonable}
\begin{equation}
\loss_{perceptual} \coloneqq  
    \sum_{l=1}^L
	\underset{j \in N^{(l)}}{\avg}
%    \frac{1}{|N^{(l)}|} \sum_{j \in N^{(l)}}
    \|\phi_{j}^{(l)}(I^S \odot \pseudoMvis) - \phi_{j}^{(l)}(I^R)\|_2^2.
\end{equation}

The visual alignment is then composed as the weighted sum over all four terms
\begin{equation}
    \loss_{visual} \coloneqq 
     \loss_{mask} + \lambda_4 \loss_{ab} +
    \lambda_5 \loss_{ms\text{-}ssim} + \lambda_6 \loss_{perceptual},
\end{equation}
with $\lambda_4, \lambda_5$ and $\lambda_6$ denoting the associated loss weights.

\myparagraph{\textbf{Geometric Alignment for Self-Supervision.}}
For geometric alignment we establish supervision leveraging the predicted pose $\predPose=[\predRot | \predTrans]$ and our pseudo pose labels $\pseudoPoseInit=[\pseudoRotInit | \pseudoTransInit]$.
Nonetheless, to better deal with noise due to severe occlusion, we do not directly employ a loss on top of $\pseudoPoseInit$ but rather use a RGB-based deep pose refiner $\Dref$~\cite{li2019deepim,labbe2020cosypose} to obtain a more robust pseudo pose label $\pseudoPose = \Dref(\pseudoPoseInit)$.
Interestingly, although the refiner is also only trained with synthetic RGB data, due to its iterative nature it can still refine $\pseudoPoseInit$ despite external variances.
We follow common procedure to utilize the point matching loss in 3D to geometrically align the 6D pose~\cite{xiang2017posecnn,li2019deepim,Wang_2021_GDRN}.
Thereby, the transformed model points of the prediction $\predPose$ are compared with the corresponding transformed model points from $\pseudoPose$ according to
\begin{equation} \label{eq:pm}
\loss_{pm} \coloneqq 
\underset{\widetilde{R} \in \mathcal{\widetilde{R}}}{\min}~
\underset{x \in \mathcal{M}}{\avg} \| (\predRot x + \predTrans) - (\widetilde{R} x + \widetilde{t}) \|_1.
\end{equation}
Notice that, to account for symmetry, $\loss_{pm}$ is computed as the minimal loss under the set of all possible known global symmetric transformations~\cite{xiang2017posecnn}.
Inspired by \Rthree{CosyPose~\cite{labbe2020cosypose}}, we also disentangle $R$, $(t_x, t_y)$ and $t_z$ in $\loss_{pm}$ following \Rthree{Simonelli \etal~\cite{Simonelli2019_disentangle_loss}}, where $t=(t_x, t_y, t_z)^T$. 
Thereby, the loss is computed for each parameter separately employing the ground truth for the remaining parameters. This leads to less noisy gradients, which improves the robustness of the optimization.

When sensor depth $D^S$ is additionally available, we can utilize it as a proxy to directly optimize the 6D pose, through geometric alignment of the target 3D model $\mathcal{M}$ against $D^S$. 
However, as the depth map only provides information for the visible areas, holistically registration \wrt the transformed 3D model harms performance. 
Therefore, we exploit the rendered depth map to enable comparison of the visible areas only. 
Nevertheless, employing a loss directly on both depth maps leads to bad correspondences as the points where the masks are not intersecting cannot be matched. 

Hence, we operate on the visible surface in 3D to find the best geometric alignment. 
We first backproject $D^S$ and $D^R$ using the corresponding masks $\pseudoMvis$ and $M^R$ to retrieve the visible pointclouds $\mathcal{P}^S$ and $\mathcal{P}^R$ in camera space with
\begin{equation} \label{eq:backproject}
\medmuskip=0mu
\thinmuskip=0mu
\thickmuskip=0mu
     \pi^{-1}(D,M,K) = 
     \Set{K^{-1}  
     \begin{pmatrix} x_j  \\ y_j \\ 1 \end{pmatrix} 
    % \begin{bmatrix} x_j  & y_j & 1 \end{bmatrix}^{T} 
     \cdot D_j} 
     {\forall j \in Pos(M)},
\end{equation}
\begin{equation}
\medmuskip=0mu
\thinmuskip=0mu
\thickmuskip=0mu
  \mathcal{P}^S \coloneqq  \pi^{-1}(D^S,\pseudoMvis,K), \quad  \mathcal{P}^R \coloneqq \pi^{-1}(D^R, M^R, K).
\end{equation}
Thereby, $(x_j,y_j)$ denotes the 2D pixel location of $j$ in the mask $M$.

Since it is infeasible to estimate direct 3D-3D correspondences between $\mathcal{P}^S$ and $\mathcal{P}^R$, we refer to the chamfer distance to seek the best alignment in 3D 
\begin{equation} \label{eq:cham}
\begin{aligned}
    \loss_{cham} \coloneqq & \underset{p^S \in \mathcal{P}^S}{\avg} 
%    \frac{1}{|\mathcal{P}^S|}\sum_{p^S \in \mathcal{P}^S}
    \min_{p^R \in \mathcal{P}^R}\|p^S - p^R\|_{2} \\ 
    + &
	\underset{p^R \in \mathcal{P}^R}{\avg} 
%    \frac{1}{|\mathcal{P}^R|}\sum_{p^R \in \mathcal{P}^R}
    \min_{p^S \in \mathcal{P}^S}\|p^S - p^R\|_{2}.
\end{aligned}
\end{equation}

The overall geometric alignment $\loss_{geom}$ is respectively defined for RGB only and RGB-D as
\begin{equation} \label{eq:geom}
\loss^{rgb}_{geom} \coloneqq \lambda_7 \loss_{pm} \quad \quad \loss^{rgb\text{-}d}_{geom} \coloneqq \lambda_7 \loss_{pm}+ \lambda_8 \loss_{cham}, 	
\end{equation}
where $\lambda_7$ and $\lambda_8$ are the corresponding loss weights.

\myparagraph{Overall Self-supervision.}
Eventually, our self-supervision $\loss_{self}$ can be summarized as a simple combination of the loss terms for visual and geometric alignment in RGB as
 \begin{equation} \label{eq:self_rgb}
     \loss^{rgb}_{self} \coloneqq \loss_{visual} + \loss^{rgb}_{geom} 
\end{equation}
or if applicable in RGB-D as
 \begin{equation} \label{eq:self_rgbd}
\loss^{rgb\text{-}d}_{self} \coloneqq \loss_{visual} + \loss^{rgb\text{-}d}_{geom}.
 \end{equation} 
\Rtwo{Noteworthy, although $\loss^{rgb\text{-}d}_{self}$ requires depth data for self-supervised training, the learned pose estimator does still not depend on depth data during inference.} Further, even when only employing RGB data alone, we can still successfully apply our self-supervision. Oppositely, the experiments from $\pselfvo$~\cite{wang2020self6d} have completely failed when discarding the depth channel.

%% file: sections/4_exp.tex
\section{Experiments}
In this section, we first describe the implementation details, employed datasets and evaluation metrics.
Afterwards, we present the analysis on the quality of predicted masks and different ablations to illustrate the effectiveness of our proposed occlusion-aware self-supervision.
We conclude by comparing our method with other state-of-the-art methods for 6D pose estimation and domain adaptation.
For better understanding, in addition to our results, we also evaluate our method using synthetic data only and additionally employing real 6D pose labels.
Since they can be considered the lower and upper bounds of our method,
we refer to them $\oursLB$ and $\oursUB$ in the following.

\subsection{Implementation Details}
\myparagraph{Training Strategy.}
We implemented our method using PyTorch~\cite{paszke2019pytorch} and ran all experiments on a NVIDIA TitanX GPU.
All networks are trained using the Ranger optimizer~\cite{liu2019radam,zhang2019lookahead,GradientCentra}.
The base learning rate is set to $10^{-4}$ and decayed after $72\%$ of the training phase using a cosine schedule~\cite{loshchilov-ICLR17SGDR}.
During pre-training on simulated data, we train the 2D object detector Yolov4~\cite{bochkovskiy2020yolov4} with a batch size of 4 for 16 epochs,
our extended GDR-Net pose estimator with a batch size of 24 for 100 epochs,
and the pose refiner with a batch size of 32 for 80 epochs.
For the deep refiner, we train our PyTorch implementation of \Rthree{DeepIM~\cite{li2019deepim}} using the publicly available synthetic data from \Rthree{BOP~\cite{hodan_bop20}} for all datasets except for YCB-Video.
We instead directly utilize the public refiner from \Rthree{CosyPose~\cite{labbe2020cosypose}} for YCB-Video, which is also pre-trained on the synthetic data from \Rthree{BOP~\cite{hodan_bop20}}.
During the self-supervised training stage, we train the pose estimator for another 100 epochs with a batch size of 6.
The teacher network is thereby updated towards the student every 10 epochs by the exponential moving average (EMA) with a momentum of 0.999~\cite{tarvainen2017mean_teacher}.
Following standard procedure~\cite{rad2017bb8,peng2019pvnet,song2020hybridpose}, we train the pose estimator and refiner separately for each object.

\myparagraph{Employed Self-supervision Hyper-Parameters.}
Our overall objective function from \eqn{eq:self_rgbd} can be broken down into the following terms {\small $\loss^{rgb\text{-}d}_{self} =
  \lambda_1 \loss_{rwce}(\pseudoMfull, M^R)
+ \lambda_2 \loss_{rwce}(\pseudoMfull, \predMfull)
+ \lambda_3 \loss_{rwce}(\pseudoMvis, \predMvis)
+ \lambda_4 \loss_{ab}
+ \lambda_5 \loss_{ms\text{-}ssim}
+ \lambda_6 \loss_{perceptual}
+ \lambda_7 \loss_{pm}
+ \lambda_8 \loss_{cham}$}.
We assigned the hyper-parameters such that their individual contributions are kept at a similar range as follows
\input{tables/loss_weights.tex}
Notice that setting $\lambda_8=0$, the self-supervision turns into our depth free formulation as denoted in \eqn{eq:self_rgb} for $\loss_{self}^{rgb}$.

\subsection{Datasets}
\subsubsection{Synthetic Training Data}
All our networks including object detector, pose estimator and pose refiner are initially pre-trained on synthetic data.
Physically-based rendering (PBR) has recently proven to be promising for improving the performance of 2D detection~\cite{hodan2019photorealistic} as well as 6D pose estimation~\cite{hodan_bop20}.
In contrast to simple OpenGL renderings~\cite{kehl2017ssd}, PBR data is more realistic and enforces physical plausibility.
Therefore, in our work we utilize the publicly available PBR data from \cite{hodan_bop20} together with various augmentations (\eg~random Gaussian noise, intensity jitter, etc.~\cite{sundermeyer2018implicit}) to train the models.
We would like to stress that pose estimators trained on synthetic data, significantly lack in performance compared to the same networks using real pose labels, even when PBR data is employed~\cite{wang2020self6d}. 
In contrast, as emphasized in Section~\ref{sec:ablation}, inferring 2D information such as visible/amodal object masks from real data is very reliable, regardless if the model was trained on synthetic data.

\subsubsection{Real-world Datasets}
To evaluate our proposed method we leverage several commonly used real-world datasets.

\myparagraph{LINEMOD~\cite{Hinterstoisser2012}}~consists of 15 sequences, each possessing $\approx$~1.2k images with clutter and lighting variations.
Only 13 of these provide \Rthree{water-tight}\footnote{\Rthree{When rendering these models artifacts can appear due to the holes within the mesh. 
In addition, the physics simulations (as done in the PBR rendering) might suffer from some undesired/unrealistic behavior.}} CAD models and we, therefore, remove the other two sequences as in other works such as \Rthree{SSD6D~\cite{kehl2017ssd}}.
Following \Rthree{Brachmann \etal~\cite{Brachmann2014Learning6O}}, we use $15\%$ of the real data for training, however, discard the accompanying pose labels.

\myparagraph{HomebrewedDB~\cite{kaskman2019homebreweddb}}~is a recently proposed dataset to evaluate the 6D pose.
We only employ the sequence which share three objects with LINEMOD~\cite{Hinterstoisser2012} to demonstrate that we can even self-supervise the same model in a different environment.

\myparagraph{Occluded LINEMOD~\cite{brachmann2014learning}}~extends one sequence of LINEMOD by additionally annotating all 8 other visible objects which often undergo severe occlusion.
\Rtwo{We adopt the BOP split~\cite{hodan2018bop} for testing and utilize the remaining samples for our self-supervised training.}

\myparagraph{YCB-Video~\cite{xiang2017posecnn}}~is a very challenging dataset consisting of 21 objects exhibiting clutter, image noise, strong occlusion and several symmetric objects.

\myparagraph{Cropped LINEMOD~\cite{bousmalis2017unsupervisedPixelda}}~is built on top of LINEMOD, including center-cropped patches of 11 different small objects in cluttered scenes portrayed in various poses.
This dataset is suitable for evaluating synthetic-to-real domain adaptation and features $\approx$~110k rendered source images, $\approx$~10k real-world target images, and $\approx$~2.6k test images from the target domain.

% ------------------------------------------------------------------------

% ------------------------------------------------------------------------
\subsection{Evaluation Metrics}
% ------------------------------------------------------------------------
We report our results referring to the common Average Distance of Distinguishable Model Points (ADD)
metric~\cite{Hinterstoisser2012}, measuring whether the average deviation of the transformed model points $e_{\tiny \text{ADD}}$ is less than $10\%$ of the object's diameter
\begin{equation} \label{eq:err_add}
    e_{\tiny \text{ADD}} = \underset{x \in \mathcal{M}}{\avg} \|(Rx + t) - ({\predRot} x + \predTrans)\|_2.
\end{equation}
For symmetric objects (\eg, Eggbox and Glue in LINEMOD) we employ the
Average Distance of Indistinguishable Model Points (ADD-S)
metric, which instead measures the error as the average distance to the closest model point~\cite{hodan2016evaluation}
\begin{equation} \label{eq:err_adds}
    e_{\tiny \text{ADD-S}} = \underset{x_{2} \in \mathcal{M}}{\avg}
    \min_{x_{1} \in \mathcal{M}} \|(Rx_{1} + t) - ({\predRot}x_{2} + \predTrans)\|_2.
\end{equation}
When evaluating on YCB-Video, we further compute the AUC (area under curve) of ADD-S/ADD(-S) by varying the distance threshold from 0cm to 10cm as in PoseCNN~\cite{xiang2017posecnn}.
Thereby, ADD-S uses the symmetric metric for all objects, while ADD(-S) only uses the symmetric metric for symmetric objects.
For Cropped LINEMOD, we report the average angle error following PixelDA~\cite{bousmalis2017unsupervisedPixelda}.

% ------------------------------------------------------------------------
\subsection{Ablation Study}
\label{sec:ablation}

% ------------------------------------------------------------------------
\myparagraph{Analysis on the Quality of Predicted Masks.}
\input{tables/table_mask_mIoU}
As self-supervision requires estimated masks of high quality from the synthetically trained model $\oursLB$,
we present quantitative results for visible and amodal masks in \tbl{tab:mask_iou}.
Thereby, we report the mIoU (\%) between the estimated masks and ground-truth masks \wrt each dataset.
Essentially, we can report an mIoU of no less than 85\% on any dataset \wrt visible masks, and more than 88\% for amodal masks referring to Occluded LINEMOD and YCB-Video.
This shows that thanks to physically-based renderings, the predicted masks from real data are very accurate, and can be therefore used as a reliable self-supervision signal. 

\input{tables/ablation_amodal_lm}
\myparagraph{\Rtwo{Ablation of Amodal Masks on LINEMOD.}} \Rtwo{Note that for datasets like LINEMOD, HomebrewedDB and Cropped LINEMOD, the occlusion is almost negligible,
hence, we only predict the visible mask and treat it as the amodal mask. 
This can be verified by the ablation study on LINEMOD as shown in \tbl{tab:ablation_amodal_lm}.
The results of predicting ``only $M_\text{vis}$'' and ``both $M_\text{amodal}$ and $M_\text{vis}$'' are almost the same for both the synthetically trained model $\oursLB$ and the self-supervised model $\ours$.}

\input{figs/2_error_loss_vs_iter}
% ------------------------------------------------------------------------
\myparagraph{Occlusion-Aware Self-Supervision \vs Pose Error.}
To demonstrate that there is indeed a high correlation between our proposed occlusion-aware self-supervision $\loss_{self}$ and the actual 6D pose error, we randomly draw 50 samples from Occluded LINEMOD and optimize separately on each sample, always initializing with $\oursLB$.
\fig{fig:error_loss_vs_iter} illustrates the average behavior \wrt loss \vs~pose error at each iteration.
As the loss decreases, also the pose error for both, rotation and translation, continuously declines until convergence. 
The accompanying qualitative images (\fig{fig:error_loss_vs_iter}, \textit{right}) further support this observation, as the initial pose is clearly worse compared to the final optimized result.

% ------------------------------------------------------------------------
\input{tables/lmo_ablation}%

\myparagraph{\Rthree{Ablation Study of Different Detectors.}}
\Rthree{In \tbl{tab:lmo_ablation}, we show that our method is robust to different detectors.
On the BOP test set of Occluded LINEMOD, by switching the synthetically trained detector from Yolov4~\cite{bochkovskiy2020yolov4} (AP: 64.3, AP50: 89.7, AP75: 74.2, Speed: 22.4~ms/img) to the slightly worse but much slower Faster R-CNN~\cite{ren2015faster} with a ResNet101~\cite{he2016deep} backbone ((AP: 62.4, AP50: 89.6, AP75: 74.0, Speed: 70.8~ms/img), there is only 1.2\% performance drop for $\ours$ -- RGB-D. 
Therefore, we use Yolov4 as the base detector in all other experiments for better accuracy and efficiency.}

\myparagraph{Effectiveness of Self-Supervision Under Occlusion.}
\tbl{tab:lmo_ablation} also illustrates the effectiveness of our self-supervision $\loss_{self}$ under occlusion referring to the BOP test set of Occluded LINEMOD.

\Rtwo{We first show that the noisy student training~\cite{Xie_2020_CVPR_noisy_student} strategy is very useful for establishing more robust self-supervision. When disabling data augmentation for the student's input, almost all objects undergo a significant performance drop, and the overall average recall drops from 64.7\% to 62.1\%.}

Note that especially geometric guidance is essential to enable self-supervision.
Disabling $\loss_{geom}$ almost always leads to instability during training with the average recall decreasing from $52.9\%$ to $5.1\%$ \wrt ADD(-S). 
Interestingly, while $\pselfvo$~\cite{wang2020self6d} also diverged when turning off $\loss_{mask}$, our new formulation exhibits more robustness which can be mostly attributed to the strong pseudo-labels produced by the teacher network together with $\Dref$.

In addition, our extension of GDR-Net to leverage amodal masks is another crucial factor for successful label-free training. 
When removing amodal mask $\Mfull$, the network suffers a significant drop from $64.7\%$ to $55.1\%$. 
While $\loss_{visual}$ and $\loss_{pm}$ only have a rather small impact, the overall best results are achieved when utilizing all loss terms together. 
Furthermore, even when only using RGB information alone, we can present great results with $59.8\%$, which is equal to an absolute improvement of $6.9\%$ over our synthetic baseline $\oursLB$. 

Most importantly, we can report a compelling improvement from $52.9\%$ to $64.7\%$ leveraging the proposed self-supervision. 
Noteworthy, our self-supervision more than halves the difference between training with and without real pose labels referring to the lower and upper bounds 
($\oursLB$ $52.9\%$ $\rightarrow$ $\ours$ $64.7\%$ $\rightarrow$ $\oursUB$ $74.4\%$).

% ------------------------------------------------------------------------
\subsection{Comparison with State of the Art}
% ------------------------------------------------------------------------
In the first part of this section we present a comparison with current state-of-the-art methods in 6D pose estimation. 
The latter part, depicts our results in the area of domain adaptation referring to Cropped LINEMOD.

\subsubsection{6D Pose Estimation}

% ------------------------------------------------------------------------
\input{tables/lm_sota}
\myparagraph{Performance on LINEMOD.}
% ------------------------------------------------------------------------
In line with other works, we distinguish between training with and without real pose labels, \ie~making use of annotated real training data.
Despite harnessing real data, we do not employ any pose labels and must, therefore, be classified as the latter. 
We want to highlight that our model can produce state-of-the-art results for training with and without labels. 
Referring to \tbl{tab:lm_sota}, for training using only synthetic data, $\oursLB$ reveals an average recall of $77.4\%$, which is deliberately better than previous state-of-the-art methods like MHP~\cite{manhardt2019ambiguity} and DPOD~\cite{zakharov2019dpod} reporting $38.8\%$ and $40.5\%$. 
On the other hand, as for training with real pose labels, we outperform all other recently published methods including PVNet~\cite{peng2019pvnet} and CDPN~\cite{li2019cdpn} reporting a mean average recall of $91.0\%$ \wrt $\oursUB$.

Notice that both of our models, RGB as well as RGB-D, come out clearly superior within all self-supervised methods reporting an average recall for $85.6\%$ and $88.5\%$ compared to $58.9\%$ for $\pselfvo$ and $60.6\%$ for Sock \etal~\cite{sock2020}. Moreover, whereas our method can be even successfully trained when depth is missing, $\pselfvo$~\cite{wang2020self6d}, in contrast, fails completely when removing the depth component, decreasing from $58.9$\% to $6.4\%$. 
We would like to stress that our proposed self-supervision is even on par with the state-of-the-art fully-supervised methods, thus, almost rendering pose labels as obsolete \wrt the LINEMOD dataset.

% ------------------------------------------------------------------------
\input{tables/hb_table}

\myparagraph{Performance on HomebrewedDB.}
% ------------------------------------------------------------------------
In ~\tbl{tab:hb}, we compare our method with DPOD~\cite{zakharov2019dpod} and SSD6D~\cite{kehl2017ssd} after refinement with~\cite{manhardt2018deep} (SSD6D+Ref.) on three objects of HomebrewedDB, which it shares with LINEMOD.
Unfortunately, methods directly solving for the 6D pose always implicitly learn the camera intrinsics which degrades the performance when exposed to a new camera.
2D-3D correspondences based approaches are instead robust to camera changes as they simply run P$n$P using the new intrinsics. 
SSD6D+Ref.~\cite{manhardt2018deep} employs contour-based pose refinement using renderings for the current hypotheses.
Similarly, rendering the pose with the new intrinsics enables again easy adaptation and can even exceed DPOD and any self-supervised approach for the Bvise object. 
In contrast, as we directly regress the pose from a single RGB image, the performance of our $\oursLB$ is worse than any other method, since we do not generalize well to the camera of HomebrewedDB. 
Nonetheless, we can still easily adapt to the new domain and intrinsics by only leveraging 15\% of unannotated data from HomebrewedDB.
In fact, we almost double the numbers for ADD for all synthetically trained methods and surpass all self-supervised approaches (\ie Self6D~\cite{wang2020self6d} and Sock \etal\cite{sock2020}) by at least by $20\%$ when only using RGB and $25\%$ when also leveraging depth data during self-supervision. 
We further almost completely close the gap of over $90\%$ between the lower and upper bounds, by pushing $\oursLB$ from $3.1\%$ to impressive $84.4\%$ compared to $\oursUB$ reporting $93.8\%$. 
Hence, our method is also very suitable for the task of domain adaption when encountering a lack of ground-truth data for the target domain.

As in our initial manuscript~\cite{wang2020self6d}, we are again curious to understand the adaptation capabilities of our model \wrt the amount of real data that we expose it to.
We divide the samples from HomebrewedDB into 100 images for testing and 900 images for training.
Afterwards, we repeatedly train our model with an increasing amount of data, however, always evaluating on the same test split.
In \fig{fig:hb} we illustrate the corresponding results.
Harnessing only as little as $10\%$ of the data for self-supervision, we can already almost achieve optimal performance of around $84\%$. 
This is a clear advantage over Self6D~\cite{wang2020self6d}, who instead requires almost $50\%$ of the data. 
In summary, we can achieve a faster adaptation speed with less data, while still easily exceeding $\pselfvo$.
\input{figs/3_hb_vary_train}

% ------------------------------------------------------------------------

\input{tables/lmo_sota}

\myparagraph{Performance on Occluded LINEMOD.}
We additionally evaluate our method on Occluded LINEMOD in \tbl{tab:lmo_sota}. 
As aforementioned, Occluded LINEMOD is a much more challenging dataset as many objects often undergo strong occlusion.
We compare the proposed methodology with state-of-the-art methods using synthetic data only under the BOP~\cite{hodan2018bop} setup. 
Thereby, our baseline approaches $\oursLB$ already clearly outperforms all other methods by a large margin. 
Exemplary, we exceed CosyPose~\cite{labbe2020cosypose}, currently top performing method from the BOP leader board~\cite{hodan_bop20}, by $6.2\%$ with $52.9\%$ compared to $46.7\%$.
Moreover, after utilizing the remaining real RGB(-D) data for self-supervision,
we again considerably enhance the performance of $\oursLB$ and again outperform both $\pselfvo$~\cite{wang2020self6d} and Sock \etal~\cite{sock2020} with a relative improvement of more than $80\%$ or respectively $100\%$ for RGB and RGB-D. 
% starting a new paragraph will raise a bug
Interestingly, despite utilizing $\Dref$~\cite{li2019deepim} in our teacher model, the performance of our RGB-D version is not limited by $\Dref$ (64.7\% \vs 62.5\%). 
However, without using depth the performance is slightly worse (59.8\% \vs 62.5\%).
Nevertheless, the iterative procedure of $\Dref$ makes it much slower than our direct regression for inference.
Concretely, given an RGB image, our pose estimator runs with $\approx$~10ms, whereas $\Dref$ needs additional 30ms on average.

\input{figs/4_lmo_qualitative}

Finally, \fig{fig:lmo_qualitative} illustrates some qualitative results on Occluded LINEMOD. 
The poses after self-supervision (\textit{green}) generally align much better with the ground-truth poses (\textit{blue}) than poses before additional self-supervision (\textit{red}). 
We would like to also point out that our self-supervised model can occasionally even produce more accurate poses than some ground-truth labels (\cf 2nd image from top left).

\input{tables/ycbv_sota}
\input{tables/ycbv_full}
\myparagraph{Performance on YCB-Video.}
In \tbl{tab:ycbv_sota} we compare our method against state of the art~\cite{xiang2017posecnn,peng2019pvnet,li2019deepim,labbe2020cosypose} on YCB-Video \wrt the common standard metric AUC of ADD-S/ADD(-S). 
In general we draw similar conclusions as for the other datasets. 
In particular, self-supervision either from RGB or RGB-D helps performance over the associated baseline. 
In addition, our approach is again almost on par with state-of-the-art methods using real pose labels for AUC of ADD(-S) with $80.0\%$ compared to $84.5\%$ from \cite{labbe2020cosypose} and even slightly better for AUC of ADD-S with $91.1\%$ \vs $90.7\%$ from $\oursUB$. 
The overall improvement from self-supervision is less significant for YBC-Video as for the other datasets and only amounts to $\approx 1\%$ or respectively $\approx 2\%$ \wrt AUC of ADD-S and AUC of ADD(-S). 
Notice that refining the predictions with $\Dref$~\cite{labbe2020cosypose} the improvements are even smaller. 
This can be mostly contributed to the fact that our baseline model is already producing very strong results for the YCB-Video without requiring real pose labels as our competitors. 
Nonetheless, we want to stress again that our self-supervision still helps when estimating the 6D pose and produces the best results for all methods that do not employ ground-truth pose labels. 
In \tbl{tab:ycb_results_AUC} we provide detailed results for each individual object.

\subsubsection{Domain Adaptation for Pose Estimation}
\input{tables/cropped_lm}
Since our method is suitable for conducting synthetic to real domain adaptation, we assess transfer skills referring to the commonly used Cropped LINEMOD scenario.
We self-supervise the model with the real training set from Cropped LINEMOD, and report the mean angle error on the real test set. 
As shown in \tbl{tab:lm_crop}, our synthetically trained model ($\oursLB$) slightly exceeds state-of-the-art methods including $\pselfvo$~\cite{wang2020self6d}.
Essentially, our approach can successfully surpass the original model on the target domain, reducing the mean angle error from $11.2\degree$ to $3.9\degree$ (or $4.7\degree$ without using depth for self-supervision).

%% file: tables/loss_weights.tex
\begin{center}
\tablestyle{4pt}{1.1}
\small
\begin{tabular}{@{}c  cccccccc@{}}
\shline
 hyper-parameter & $\lambda_1$ & $\lambda_2$ & $\lambda_3$ & $\lambda_4$ & $\lambda_5$ & $\lambda_6$ & $\lambda_7$ & $\lambda_8$ \\
% \shline
\hline
 value & 1        & 1        & 1         & 0.2         & 1         & 0.15        & 10          & \{100, 0\} \\
% \hline
\shline
\end{tabular}
\end{center}

%% file: tables/table_mask_mIoU.tex
\begin{table}[t!]
\centering
\caption{\label{tab:mask_iou}
    Results of mask quality on each dataset \wrt mIoU (\%)
}
\begin{threeparttable}[c]
\tablestyle{18pt}{1.3}
\begin{tabular}{@{}l  cc @{}}
% \shline
% \hline
\shline
Mask type                                                & Visible & Amodal \\
% \shline
\hline
LINEMOD~\cite{Hinterstoisser2012} $^\dagger$                       & 93.8 & - \\
HomebrewedDB~\cite{kaskman2019homebreweddb} $^\dagger$              & 93.5 & - \\
Cropped LINEMOD~\cite{bousmalis2017unsupervisedPixelda} $^\dagger$  & 89.3 & - \\
% \hline
Occluded LINEMOD~\cite{brachmann2014learning}            & 85.3 & 88.8 \\
YCB-Video~\cite{xiang2017posecnn}                        & 88.8 & 92.1 \\
% \hline
\shline
\end{tabular}
\begin{tablenotes}
    \item[] \Rtwo{$^\dagger$ For datasets with almost negligible occlusions, we only predict the visible mask and treat it as the amodal mask.}
\end{tablenotes}
\vspace{-0.26em}
\end{threeparttable}
\end{table}

%% file: tables/ablation_amodal_lm.tex
\begin{table}[t]
\caption{\label{tab:ablation_amodal_lm}
\Rtwo{Ablation study of amodal masks on LINEMOD using \\
the Average Recall (\%) of ADD(-S) metric}
}
\centering
% \small
\tablestyle{24pt}{1.3}
\begin{tabular}{@{}l c c c@{}}
% \hline 
\shline
          & $M_\text{amodal}$ & $M_\text{vis}$ & ADD(-S) \\
% \shline
\hline
$\oursLB$ & \cmark & \cmark & 77.2 \\
$\oursLB$ &  & \cmark  & 77.4 \\
% \hline
$\ours$   & \cmark & \cmark & 88.3 \\
$\ours$   & & \cmark & 88.5 \\
% \hline 
\shline
\end{tabular}
\end{table}

%% file: figs/2_error_loss_vs_iter.tex
\begin{figure}[t!]
\centering
	\includegraphics[width=1\linewidth]{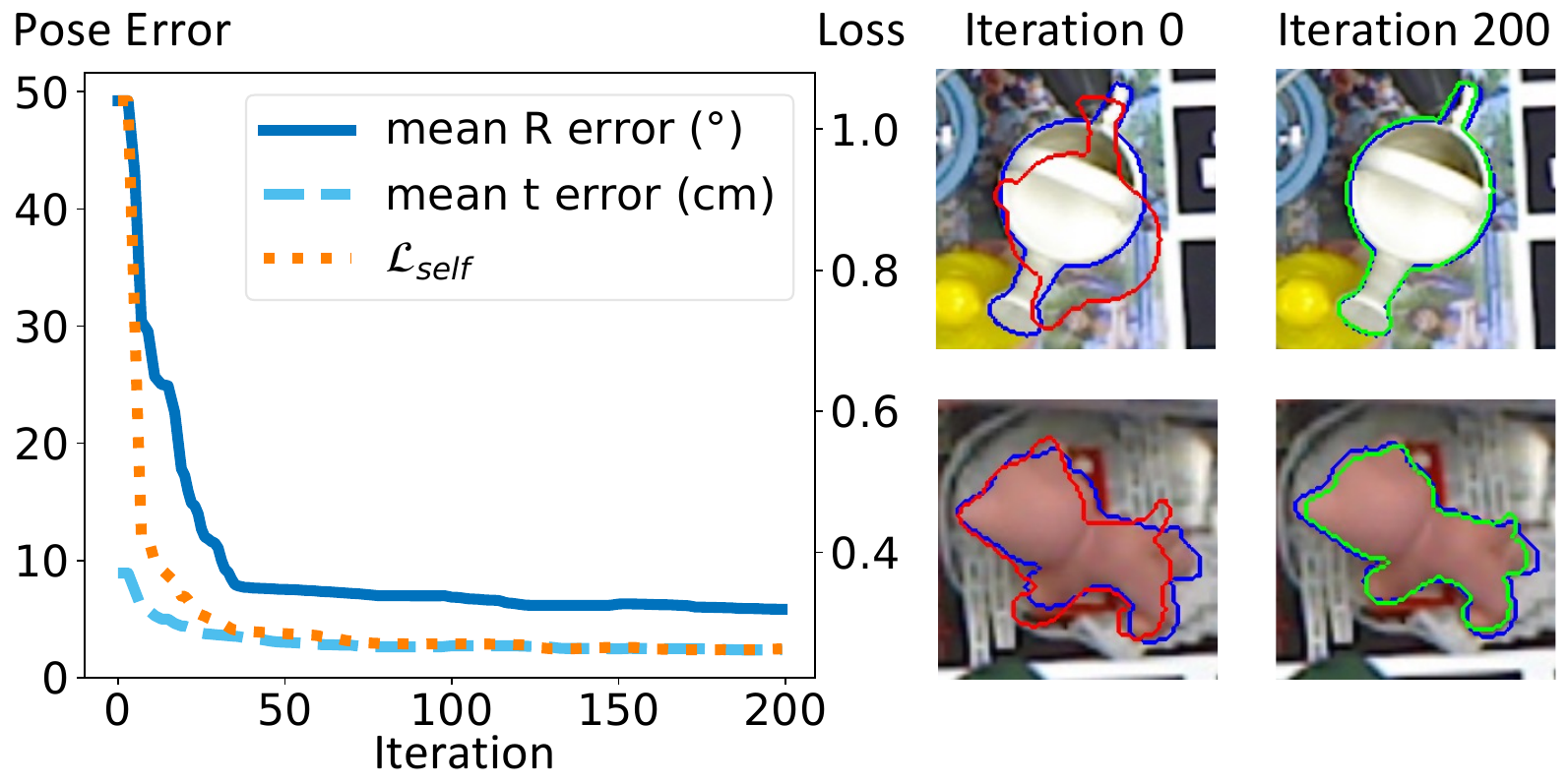}
	\vspace{-2em}
\caption{\label{fig:error_loss_vs_iter}
	\emph{Occlusion-aware self-supervision \vs pose error.}
	\emph{Left:} We optimize $\loss_{self}$ on single images from Occluded LINEMOD for 200 iterations and report the average over in total 50 images. We initialize the 6D poses with $\oursLB$.
	\emph{Right:} Visualization of the ground-truth pose (\emph{blue}) and predicted poses at iteration 0 (\emph{red}) and 200 (\emph{green}).
}
\end{figure}

%% file: tables/lmo_ablation.tex
\begin{table*}
\centering
\caption{\label{tab:lmo_ablation}
    Ablation study on \Rtwo{the BOP test set} of Occluded LINEMOD \wrt the Average Recall (\%) of ADD(-S)
}
\begin{threeparttable}[c]
\tablestyle{12pt}{1.3}
\begin{tabular}{@{}>{\columncolor{white}[0pt][\tabcolsep]}l c c c c c c c c >{\columncolor{white}[\tabcolsep][0pt]}c@{}}
% \hline 
\shline
Method & Ape  & Can & Cat & Drill & Duck  & Eggbox$^*$ & Glue$^*$  & Holep  & Mean \\
% \shline
\hline
\rowcolor{RthreeColorBG}\multicolumn{10}{c}{Detector: Faster R-CNN~\cite{ren2015faster} trained with synthetic data (AP: 62.4, AP50: 89.6, AP75: 74.0, Speed: 70.8~ms/img)} \\
% \hline
\rowcolor{RthreeColorBG} $\oursLB$                  & 44.6 & 86.9 & 48.0 & 89.5 & 12.8 & 35.0 & 72.9 & 35.0 & 53.1 \\
\rowcolor{RthreeColorBG} $\oursLB$ + $\Dref$        & 52.6 & \winAllwoPose{97.5} & 56.7 & \winAllwoPose{97.5} & 27.2 & \winNoPose{56.1} & 88.6 & 23.5 & 62.5 \\
\rowcolor{RthreeColorBG} $\ours$ -- RGB-D           & 58.3 & 95.0 & 56.7 & 92.0 & 31.1 & 55.0 & 87.1 & 32.5 & 63.5 \\
%%%%%%%%%%%%%%%%%%%%%%%%%%%%%%%%%%%%%%%%%%%%%%%%%%%%%%%%%%%%%%%%%%%%%
% \shline
\rowcolor{RthreeColorBG}\multicolumn{10}{c}{Detector: Yolov4~\cite{bochkovskiy2020yolov4} trained with synthetic data (AP: 64.3, AP50: 89.7, AP75: 74.2, Speed: 22.4~ms/img)} \\
% \hline
$\oursLB$                  & 44.0 & 83.9 & 49.1 & 88.5 & 15.0 & 33.9 & 75.0 & 34.0 & 52.9 \\
$\oursLB$ + $\Dref$ & 53.7 & 97.0 & 60.2 & 96.5 & 27.8 & 51.1 & \winAllwoPose{89.3} & 24.5 & 62.5 \\
% \hline 
$\oursUB$ & 53.7 & 93.5 & 57.3 & 91.5 & \winAll{71.7} & \winAll{57.8} & 88.6 & \winAll{81.0} & \winAll{74.4}\\
% \hline 
\rowcolor{RtwoColorBG}w/o Input Augmentation& 56.6 & 94.0 & 52.6 & 92.0 & 29.4 & 50.0 & 87.9 & 34.0 & 62.1 \\
w/o $\loss_{geom}$    & 0.0 & 0.0 & 0.0 & 19.5 & 0.0 & 7.2 & 1.4 & 13.0 & 5.1 \\
w/o $M_\text{amodal}$ & 44.0 & 84.9 & 53.8 & 87.5 & 14.4 & 39.4 & 82.1 & 34.5 & 55.1  \\
w/o $\loss_{visual}$  & 58.9 & 95.5 & 56.7 & 93.0 & 32.2 & 48.3 & 87.9 & 33.5 & 63.2  \\
w/o $\loss_{pm}$      & 51.4 & 88.9 & 46.8 & 90.5 & 41.7 & 51.7 & \winAllwoPose{89.3} & \winNoPose{54.0} & 64.3  \\
% \hline 
$\ours$ -- RGB (w/o $\loss_{cham}$) & 57.7 & 95.0 & 52.6 & 90.5 & 26.7 & 45.0 & 87.1 & 23.5 & 59.8  \\
% \hline
$\ours$ -- RGB-D & \winAllwoPose{59.4} & 96.5 & \winAllwoPose{60.8} & 92.0 & \winNoPose{30.6} & 51.1 & 88.6 & 38.5 & \winNoPose{64.7} \\
%%%%%%%%%%%%%%%%%%%%%%%%%%%%%%%%%%%%%%%%%%%%%%%%%%%%%%%%%%%%%%%%%%%%%
% \hline
\shline 
\end{tabular}
\begin{tablenotes}
    \item[] $^*$ denotes symmetric objects;
    \item[] $^\dagger$ The best label-free method is marked in bold, and the overall best method is underlined.
\end{tablenotes}
\end{threeparttable}
\end{table*}

%% file: tables/lm_sota.tex
\begin{table*}[t]
\caption{\label{tab:lm_sota}
Results on LINEMOD referring to the Average Recall (\%) of ADD(-S) metric
% note that Self6D++(UB)'s object detector is trained with synthetic data only, so it is slightly worse than original GDR-Net
}
\centering
\scalebox{0.95}{
\begin{threeparttable}[c]
\tablestyle{1pt}{1.3}
\begin{tabular}{@{}l >{\columncolor{RthreeColorBG}}c >{\columncolor{RthreeColorBG}}c  ccccccccccccc c@{}}
% \hline 
\shline
\multirow{2}{*}{Method} & \#Params & Synthetic &  \multicolumn{13}{c}{Object} & \multirow{2}{*}{Mean}\\
% \cmidrule{2-17}
\cline{4-16}
       & (M) & Data & Ape & Bvise & Cam & Can & Cat & Drill & Duck  & Eggbox$^*$ & Glue$^*$  & Holep & Iron  & Lamp  & Phone &  \\
\hline
% syn methods
\multicolumn{17}{c}{Supervision: Syn} \\
AAE~\cite{sundermeyer2018implicit} &$36.5^\text{(d)}$+$29.7^\text{(p)}\x13$ & OpenGL  & 4.0  & 20.9 & 30.5 & 35.9 & 17.9 & 24.0 & 4.9  & 81.0  & 45.5 & 17.6 & 32.0 & 60.5 & 33.8 & 31.4 \\
MHP~\cite{manhardt2019ambiguity} & $92.1^\text{(p)}\x12$ & OpenGL & 11.9 & 66.2 & 22.4 & 59.8 & 26.9 & 44.6 & 8.3 & 55.7 & 54.6 & 15.5 & 60.8 & - & 34.4 & 38.8 \\
DPOD~\cite{zakharov2019dpod} $^\dagger$& $14.0^\text{(p)}\x13$ & OpenGL & 35.1 & 59.4 & 15.5 & 48.8 & 28.1 & 59.3 & 25.6 & 51.2 & 34.6 & 17.7 & 84.7 & 45.0 & 20.9 & 40.5 \\
\rowcolor{RthreeColorBG}DPOD+Ref.~\cite{zakharov2019dpod} $^\dagger$ & $(14.0^\text{(p)}$+$16.5^\text{(r)})\x13$ & OpenGL & 52.1 &	64.7 & 22.2 &	77.5 & 56.5 & 65.2 & 49.0 & 62.2 & 38.9 & 25.6 & 98.4 & 58.4 & 33.8 & 54.2 \\
$\oursLB$                         & $96.5^\text{(d)}$+$42.7^\text{(p)}\x13$ & PBR & 50.9 & \winNoPose{99.4} & 89.2 & 97.2 & 79.9 & 98.7 & 24.6 & 81.1  & 81.2 & \winNoPose{41.9} & 98.8 & 98.9 & 64.3 & 77.4 \\
$\oursLB$+$\Dref$  & $96.5^\text{(d)}$+$(42.7^\text{(p)}$+$37.6^\text{(r)})\x13$  & PBR & \winAllwoPose{85.8} & 93.1 & \winAllwoPose{99.1} & \winAllwoPose{99.8} & \winNoPose{91.5} & \winAllwoPose{100.0} & 61.9 & 93.5 & 93.3 & 32.1 & \winAllwoPose{100.0} & \winNoPose{99.1} & \winAllwoPose{94.8} & 88.0 \\
% with real labels
% \hline 
\multicolumn{17}{c}{Supervision: Syn + Real GT} \\
YOLO6D~\cite{tekin18_yolo6d}& $50.5^\text{(p)}\x13$ & \xmark & 21.6 & 81.8 & 36.6 & 68.8 & 41.8 & 63.5 & 27.2 & 69.6 & 80.0 & 42.6 & 75.0 & 71.1 & 47.7 & 56.0 \\
DPOD~\cite{zakharov2019dpod}  $^\dagger$& $14.0^\text{(p)}\x13$& OpenGL & 53.3 & 95.2 & 90.0 & 94.1 & 60.4 & 97.4 & 66.0 & 99.6 & 93.8 & 64.9 & 99.8 & 88.1 & 71.4 & 82.6 \\
PVNet~\cite{peng2019pvnet} & $13.0^\text{(p)}\x13$ & Blender & 43.6 & \winAll{99.9} & 86.9 & 95.5 & 79.3 & 96.4 & 52.6 & 99.2 & 95.7 & 81.9 & 98.9 & 99.3 & 92.4 & 86.3 \\
CDPN~\cite{li2019cdpn} & $41.4^\text{(d)}$+$113.5^\text{(p)}$ & OpenGL & 64.4 & 97.8 & 91.7 & 95.9 & 83.8 & 96.2 & 66.8 & 99.7 & \winAll{99.6} & \winAll{85.8} & 97.9 & 97.9 & 90.8 & 89.9 \\
$\oursUB$  & $96.5^\text{(d)}$+$42.7^\text{(p)}\x13$ & PBR & 85.0 & 99.8 & 96.5 & 99.3 & \winAll{93.0} & \winAll{100.0} & 65.3 & \winAll{99.9} & 98.1 & 73.4 & 86.9 & \winAll{99.6} & 86.3 & \winAll{91.0} \\
% self methods
% \hline 
\multicolumn{17}{c}{Supervision: Syn + Self} \\
$\pselfvo$~\cite{wang2020self6d} w/o Depth & $155.1^\text{(p)}$ & PBR+OpenGL & 0.0 & 10.1 & 3.1 & 0.0 & 0.0 & 7.5 & 0.1 & 33.0 & 0.2 & 0.0 & 5.9 & 20.7 & 2.4 & 6.4 \\
$\pselfvo$~\cite{wang2020self6d} & $155.1^\text{(p)}$ & PBR+OpenGL & 38.9 & 75.2 & 36.9 & 65.6 & 57.9 & 67.0 & 19.6 & \winNoPose{99.0} & 94.1 & 16.2 & 77.9 & 68.2 & 50.1 & 58.9 \\
Sock \etal~\cite{sock2020}& $60.3^\text{(d)}$+$25.7^\text{(p)}\x13$ & NMD~\cite{kato2018renderer} & 37.6 & 78.6 & 65.5 & 65.6 & 52.5 & 48.8 & 35.1 & 89.2 & 64.5 & 41.5 & 80.9 & 70.7 & 60.5 & 60.6 \\
$\ours$ -- RGB & $96.5^\text{(d)}$+$42.7^\text{(p)}\x13$ & PBR & 76.0 & 91.6 & 97.1 & \winAllwoPose{99.8} & 85.6 & 98.8 & 56.5 & 91.0 & 92.2 & 35.4 & 99.5 & 97.4 & 91.8 & 85.6 \\
$\ours$ -- RGB-D & $96.5^\text{(d)}$+$42.7^\text{(p)}\x13$ & PBR & 75.4 & 94.9 & 97.0 & 99.5 & 86.6 & 98.9 & \winAllwoPose{68.3} & \winNoPose{99.0} & \winNoPose{96.1} & \winNoPose{41.9} & 99.4 & 98.9 & 94.3 & \winNoPose{88.5} \\
\shline
\end{tabular}
\begin{tablenotes}
    \item[] $^*$ denotes symmetric objects;
    \Rthree{$^\dagger$ The numbers of DPOD~\cite{zakharov2019dpod} are different from those in their paper since they used average precision instead. The authors provided us with their results for average recall;}
    \item[] $^\ddagger$ The best label-free method is marked in bold, and the overall best method is underlined;
    \Rthree{$^\text{(d) (p) (r)}$ respectively denotes the detector, pose estimator, and refiner.}
\end{tablenotes}
\end{threeparttable}
}
\end{table*}

%% file: tables/hb_table.tex
\begin{table*}[t]
\centering
\caption[]{\label{tab:hb}
    Results on HomebrewedDB
}
\scalebox{0.93}{
\begin{threeparttable}[c]
\tablestyle{22pt}{1.3}
\begin{tabular}{
@{}
    l >{\columncolor{RthreeColorBG}}c >{\columncolor{RthreeColorBG}}c ccc c
@{}
}
% \hline 
\shline
\multirow{2}{*}{Method} & \#Params & Synthetic & \multicolumn{3}{c}{Object} & \multirow{2}{*}{Mean} \\
\cmidrule(lr){4-6} 
 & (M) & Data &  Bvise & Drill & Phone &  \\
% \shline 
\hline
\multicolumn{7}{c}{Supervision: Syn} \\
$\oursLB$ & $96.5^\text{(d)}$+$42.7^\text{(p)}\x3$ & PBR & 7.1 & 2.2 & 0.1 & 3.1 \\
DPOD~\cite{zakharov2019dpod} $^\dagger$ & $14.0^\text{(p)}\x3$  & OpenGL & 52.9 & 37.8 & 7.3 & 32.7 \\
SSD6D+Ref. \cite{manhardt2018deep} $^\dagger$ & $(92.1^\text{(p)}$+$5.4^\text{(r)})\x3$ & OpenGL & \winNoPose{82.0} & 22.9 & 24.9 & 43.3 \\
% \hline 
\multicolumn{7}{c}{Supervision: Syn + Real GT} \\
$\oursUB$ & $96.5^\text{(d)}$+$42.7^\text{(p)}\x3$ & PBR & \winAll{98.6} & 97.7 & 85.1 & \winAll{93.8}  \\ 
% \hline 
\multicolumn{7}{c}{Supervision: Syn + Self} \\
Sock \etal~\cite{sock2020}       & $60.3^\text{(d)}$+$25.7^\text{(p)}\x3$ & NMD~\cite{kato2018renderer} & 57.3 & 46.6 & 41.5 & 52.0 \\
$\pselfvo$~\cite{wang2020self6d} & $155.1^\text{(p)}$ & PBR+OpenGL & 72.1 & 65.1 & 41.8 & 59.7 \\
$\ours$ -- RGB   & $96.5^\text{(d)}$+$42.7^\text{(p)}\x3$ & PBR & 56.1 & 97.7 & 85.1 & 79.6   \\
$\ours$ -- RGB-D & $96.5^\text{(d)}$+$42.7^\text{(p)}\x3$ & PBR & 67.1 & \winAllwoPose{98.0} & \winAllwoPose{88.2} & \winNoPose{84.4}  \\
% \hline 
\shline
\end{tabular}
\begin{tablenotes}
    \item[] \Rthree{$^\dagger$ The numbers are different as in their paper since they used average precision instead. The authors provided us with their results for average recall;}
    \item[] $^\ddagger$ The best label-free method is in bold, and the overall best method is underlined;
    \Rthree{$^\text{(d) (p) (r)}$ respectively denotes the detector, pose estimator, and refiner.}
\end{tablenotes}
\end{threeparttable}
}
\end{table*}

%% file: figs/3_hb_vary_train.tex
\begin{figure}[b]
\centering
  \includegraphics[width=0.86\linewidth,valign=m]{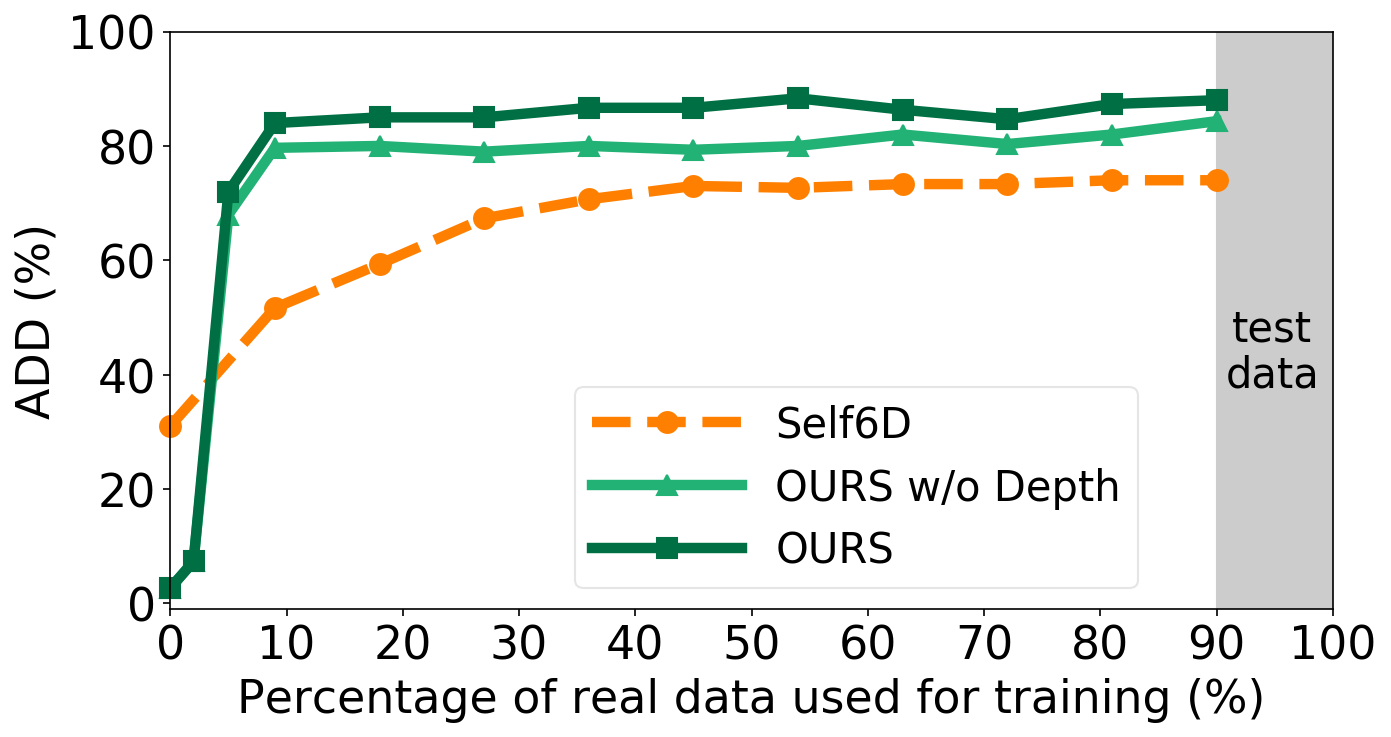}
\caption{\label{fig:hb}
  Self-supervised training \wrt an increasing percentage of real training data on HomebrewedDB. Results are always reported on the same unseen test split.
}
\end{figure}

%% file: tables/lmo_sota.tex
\begin{table*}[t!]
\centering
\caption[]{\label{tab:lmo_sota}
     Results on \Rtwo{the BOP test set} of Occluded LINEMOD \wrt the Average Recall (\%) of ADD(-S)
} 
\begin{threeparttable}[c]
\tablestyle{6pt}{1.3}
\begin{tabular}{@{}l >{\columncolor{RthreeColorBG}}c >{\columncolor{RthreeColorBG}}c cccccccc c@{}}
% \hline  
\shline
\multirow{2}{*}{Method} & \#Params & Synthetic & \multicolumn{8}{c}{Object} & \multirow{2}{*}{Mean}\\
% \cline{4-11}
\cline{4-11}
& (M)  & Data & Ape  & Can & Cat & Drill & Duck  & Eggbox$^*$ & Glue$^*$  & Holep &  \\
% \shline
\hline
\multicolumn{12}{c}{Supervision: Syn} \\
DPOD~\cite{zakharov2019dpod} $^\dagger$      & $14.0^\text{(p)}\x8$ & OpenGL & 2.3 & 4.0 & 1.2 & 10.5 & 7.2  & 4.4 &  12.9 & 7.5 & 6.3 \\
CDPN~\cite{li2019cdpn} $^\dagger$            & $55.4^\text{(d)}$+$13.0^\text{(p)}\x8$  & Blender & 20.0 & 15.1 & 16.4 & 5.0 & 22.2 & 36.1 & 27.9 & 24.0 & 20.8 \\
CDPNv2~\cite{li2019cdpn} $^\dagger$          & $49.7^\text{(d)}$+$26.1^\text{(p)}\x8$  & PBR & 20.6 & 64.8 & 24.0 & 60.0 & 42.2 & 40.0 & 66.4 & \winNoPose{42.0} & 45.0 \\
CosyPose~\cite{labbe2020cosypose} $^\dagger$ & $44.0^\text{(d)}$+$10.7^\text{(p)}$+$10.7^\text{(r)}$ & PBR & 44.0 & 69.9 & 42.1 & 67.5 & \winNoPose{47.8} & 24.4 & 60.0 & 17.5 & 46.7 \\
$\oursLB$ & $96.5^\text{(d)}$+$42.7^\text{(p)}\x8$ & PBR & 44.0 & 83.9 & 49.1 & 88.5 & 15.0 & 33.9 & 75.0 & 34.0 & 52.9 \\
$\oursLB$+$\Dref$  & $96.5^\text{(d)}$+$(42.7^\text{(p)}$+$37.6^\text{(r)})\x8$  & PBR & 53.7 & \winAllwoPose{97.0} & 60.2 & \winAllwoPose{96.5} & 27.8 & 51.1 & \winAllwoPose{89.3} & 24.5 & 62.5 \\    
% \hline 
\multicolumn{12}{c}{Supervision: Syn + Real GT} \\
$\oursUB$ & $96.5^\text{(d)}$+$42.7^\text{(p)}\x8$ & PBR &  53.7 & 93.5 & 57.3 & 91.5 & \winAll{71.7} & \winAll{57.8} & 88.6 & \winAll{81.0} & \winAll{74.4}\\
% \hline 
\multicolumn{12}{c}{Supervision: Syn + Self} \\
Sock \etal~\cite{sock2020} & $60.3^\text{(d)}$+$25.7^\text{(p)}\x8$ & NMD~\cite{kato2018renderer} & 12.0 & 27.5 & 12.0 & 20.5 & 23.0 & 25.1 & 27.0 & 35.0 & 22.8 \\
$\pselfvo$~\cite{wang2020self6d} &$155.1^\text{(p)}$ & PBR+OpenGL & 13.7 & 43.2 & 18.7 & 32.5 & 14.4 & \winAllwoPose{57.8} & 54.3 & 22.0 & 32.1 \\
$\ours$ -- RGB & $96.5^\text{(d)}$+$42.7^\text{(p)}\x8$ & PBR & 57.7 & 95.0 & 52.6 & 90.5 & 26.7 & 45.0 & 87.1 & 23.5 & 59.8  \\
$\ours$ -- RGB-D & $96.5^\text{(d)}$+$42.7^\text{(p)}\x8$ & PBR & \winAllwoPose{59.4} & 96.5 & \winAllwoPose{60.8} & 92.0 & 30.6 & 51.1 & 88.6 & 38.5 & \winNoPose{64.7} \\
\shline 
\end{tabular}
\begin{tablenotes}
    \item[] $^*$ denotes symmetric objects;
    \item[] $^\dagger$ The results are re-evaluated with ADD(-S) metric using the estimated poses for the BOP 2019 and 2020 challenges~\cite{hodan_bop20};
    \item[] $^\ddagger$ The best pose label free method is marked in bold, and the overall best method is underlined;
    \Rthree{$^\text{(d) (p) (r)}$ respectively denotes the detector, pose estimator, and refiner.}
\end{tablenotes}
\end{threeparttable}
% \vspace{-2mm}
\end{table*}

%% file: figs/4_lmo_qualitative.tex
\begin{figure*}[t!]
\centering
\includegraphics[width=0.95\textwidth]{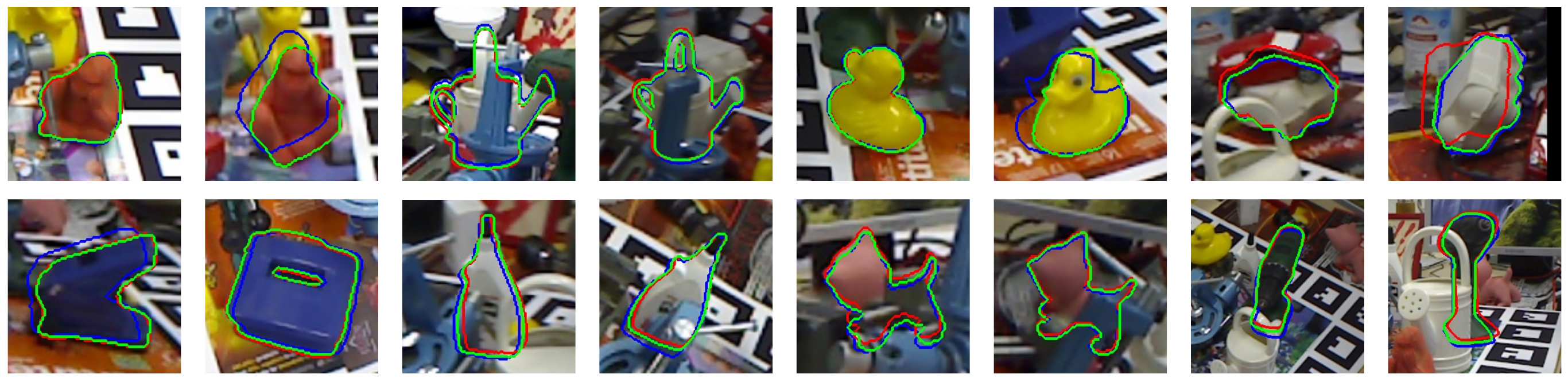}
\caption{
Qualitative results on Occluded LINEMOD.
The \textit{Blue}, \textit{Red} and \textit{Green} silhouettes represent the ground-truth 6D pose, the results before and after applying our self-supervision, respectively.
}
\label{fig:lmo_qualitative}
\end{figure*}

%% file: tables/ycbv_sota.tex
\begin{table*}[t!]
\centering
\caption[]{\label{tab:ycbv_sota}
    Results on YCB-Video using AUC of ADD-S/ADD(-S) metrics
} 
\begin{threeparttable}[c]
\tablestyle{30pt}{1.3}
\begin{tabular}{@{}l >{\columncolor{RthreeColorBG}}c >{\columncolor{RthreeColorBG}}c cc@{}}
% \hline  
\shline
\multirow{2}{*}{Method} & \#Params & Synthetic & AUC of & AUC of \\
& (M) & Data & ADD-S & ADD(-S) \\
% \shline 
\hline 
\multicolumn{5}{c}{Supervision: Syn} \\
$\oursLB$  &   $96.5^\text{(d)}$+$42.7^\text{(p)}\x21$       &   PBR         & 89.4 & 77.8 \\
$\oursLB$+$\Dref$ $^\dagger$ & $96.5^\text{(d)}$+$42.7^\text{(p)}\x21$+$10.7^\text{(r)}$& PBR & 90.1 & 79.2 \\
% \hline
\multicolumn{5}{c}{Supervision: Syn + Real GT} \\
PoseCNN~\cite{xiang2017posecnn}   & $135.2^\text{(p)}$ & Blender & 75.9 & 61.3 \\
PVNet~\cite{peng2019pvnet}        & $13.0^\text{(p)}\x21$ & Blender  & -    & 73.4 \\
DeepIM~\cite{li2019deepim}        & $135.2^\text{(p)}$+$37.6^\text{(r)}$    & OpenGL & 88.1 & 81.9 \\
CosyPose~\cite{labbe2020cosypose} & $135.2^\text{(d)}$+$10.7^\text{(p)}$+$10.7^\text{(r)}$ & OpenGL & 89.8 & \winAll{84.5} \\
$\oursUB$                         & $96.5^\text{(d)}$+$42.7^\text{(p)}\x21$ & PBR & 90.7 & 82.6 \\
% \hline 
\multicolumn{5}{c}{Supervision: Syn + Self} \\
$\ours$ -- RGB   & $96.5^\text{(d)}$+$42.7^\text{(p)}\x21$  & PBR & 90.5 & 78.9 \\
$\ours$ -- RGB-D   & $96.5^\text{(d)}$+$42.7^\text{(p)}\x21$ & PBR & \winAllwoPose{91.1} & \winNoPose{80.0} \\
% \hline 
\shline
\end{tabular}
\begin{tablenotes} 
    \item[] $^\dagger$ We employ the publicly available synthetically trained refiner from CosyPose~\cite{labbe2020cosypose} as $\Dref$;
    \item[] $^\ddag$ The best pose label-free method is marked in bold, and the overall best method is underlined; 
    \item[] `-' denotes unavailable results;
    \Rthree{$^\text{(d) (p) (r)}$ respectively denotes the detector, pose estimator, and refiner.}
\end{tablenotes}
\end{threeparttable}
\end{table*}

%% file: tables/ycbv_full.tex
\begin{table*}
\centering
\caption{\label{tab:ycb_results_AUC}
    Detailed results on YCB-Video \wrt AUC of ADD-S and ADD(-S)
}
\begin{threeparttable}[c]
\tablestyle{4.0pt}{1.3}
\begin{tabular}{l cc cc cc cc cc}
\shline  
 Supervision  & \multicolumn{4}{c}{Syn} & \multicolumn{4}{c}{Syn + Self} & \multicolumn{2}{c}{Syn + Real GT}  \\
% \midrule
% \hline 
\cline{2-11}
 Method  &
 \multicolumn{2}{c}{\tabincell{c}{$\oursLB$}} & \multicolumn{2}{c}{\tabincell{c}{$\oursLB$+$D_\text{Ref}$}} &
 \multicolumn{2}{c}{\tabincell{c}{$\ours$ -- RGB}} & \multicolumn{2}{c}{\tabincell{c}{$\ours$ -- RGB-D}} &
 \multicolumn{2}{c}{\tabincell{c}{$\oursUB$}}  \\
\cline{2-11}
Metric & \tabincell{c}{AUC of\\ADD-S} & \tabincell{c}{AUC of \\ ADD(-S)}
                  & \tabincell{c}{AUC of\\ADD-S} & \tabincell{c}{AUC of \\ ADD(-S)}
                  & \tabincell{c}{AUC of\\ADD-S} & \tabincell{c}{AUC of \\ ADD(-S)}
                  & \tabincell{c}{AUC of\\ADD-S} & \tabincell{c}{AUC of \\ ADD(-S)}
                  & \tabincell{c}{AUC of\\ADD-S} & \tabincell{c}{AUC of \\ ADD(-S)}\\
% \shline 
\hline  
002\_master\_chef\_can        & \winNoPose{89.5} & \winNoPose{9.6} & 81.8 & 8.0   & 86.4  & 8.0  & 88.8 & 8.4  & \winAll{93.8} & \winAll{56.7}  \\
003\_cracker\_box             & 94.5 & 85.6 & \winNoPose{95.4} & \winNoPose{87.1} & 93.4  & 83.6 & 94.2 & 84.9 & \winAll{98.8} & \winAll{92.8}  \\
004\_sugar\_box               & 93.2 & 82.4 & \winNoPose{95.9} & 87.6 & 95.1  & 86.5 & 95.8 & \winNoPose{88.0} & \winAll{99.6} & \winAll{95.0}  \\
005\_tomato\_soup\_can        & 89.8 & 77.5 & \winNoPose{94.4} & \winNoPose{84.3} & 89.6  & 77.3 & 90.8 & 79.4 & \winAll{95.4} & \winAll{90.5}  \\
006\_mustard\_bottle          & 97.5 & 92.0 & 96.7 & 88.7 & \winNoPose{98.9}  & 91.7 & 98.6 & \winNoPose{92.7} & \winAll{100.0} & \winAll{94.7}  \\
007\_tuna\_fish\_can          & 96.0 & 86.7 & 95.4 & 85.4 & 96.3  & 86.7 & \winNoPose{97.5} & \winNoPose{89.7} & \winAll{99.9} & \winAll{97.0}  \\
008\_pudding\_box             & 96.1 & 89.7 & 84.5 & 71.7 & 96.5  & 89.7 & \winAllwoPose{98.4} & \winAllwoPose{93.9} & 63.3 & 42.1  \\
009\_gelatin\_box             & 90.2 & 78.9 & \winAllwoPose{94.4} & \winAllwoPose{86.4} & 91.4  & 80.4 & 94.0 & 83.9 & 92.9 & 84.7  \\
010\_potted\_meat\_can        & \winNoPose{90.4} & 74.7 & 90.1 & 72.6 & 88.2  & 74.9 & 89.3 & \winNoPose{75.7} & \winAll{91.1} & \winAll{78.2}  \\
011\_banana                   & \winAllwoPose{99.2} & \winAllwoPose{92.9} & 97.6 & 90.4 & 97.5  & 91.4 & 98.5 & 91.8 & 93.0 & 80.5  \\
019\_pitcher\_base            & 97.8 & 89.2 & \winAllwoPose{99.5} & \winNoPose{94.7} & 98.7  & 89.9 & 98.9 & 92.1 & 99.3 & \winAll{98.7}  \\
021\_bleach\_cleanser         & 90.5 & 80.3 & 87.9 & 76.8 & 91.9  & 81.7 & \winAllwoPose{93.5} & \winAllwoPose{84.5} & 91.2 & 81.9  \\
024\_bowl$^*$                 & 77.6 & 77.6 & 88.3 & 88.3 & 89.0  & 89.0 & \winAllwoPose{89.1} & \winAllwoPose{89.1} & 87.2 & 87.2  \\
025\_mug                      & 90.1 & 73.1 & 93.9 & 81.0 & 91.8  & 77.4 & \winNoPose{94.1} & \winNoPose{81.4} & \winAll{96.4} & \winAll{86.6}  \\
035\_power\_drill             & 94.7 & \winNoPose{84.6} & 94.7 & 83.4 & 95.1  & 83.6 & \winNoPose{95.2} & 84.2 & \winAll{99.7} & \winAll{93.6}  \\
036\_wood\_block$^*$          & 76.8 & 76.8 & 59.8 & 59.8 & 77.2  & 77.2 & \winAllwoPose{78.3} & \winAllwoPose{78.3} & 68.6 & 68.6  \\
037\_scissors                 & 74.1 & 55.6 & \winAllwoPose{89.2} & \winAllwoPose{75.8} & 68.2  & 45.5 & 69.2 & 45.2 & 78.9 & 61.3  \\
040\_large\_marker            & 82.9 & 70.5 & 86.5 & 74.8 & 87.3  & \winNoPose{75.3} & \winNoPose{87.5} & 74.6 & \winAll{93.0} & \winAll{81.7}  \\
051\_large\_clamp$^*$         & 76.5 & 76.5 & 82.8 & 82.8 & \winAllwoPose{83.8}  & \winAllwoPose{83.8} & 79.2 & 79.2 & 81.7 & 81.7  \\
052\_extra\_large\_clamp$^*$  & 84.6 & 84.6 & 87.1 & 87.1 & 87.1  & 87.1 & \winAllwoPose{87.3} & \winAllwoPose{87.3} & 86.9 & 86.9  \\
061\_foam\_brick$^*$          & 94.4 & 94.4 & 95.6 & 95.6 & \winAllwoPose{96.8}  & \winAllwoPose{96.8} & 95.5 & 95.5 & 94.3 & 94.3  \\
% \midrule
% \hline 
Mean                          & 89.4 & 77.8 & 90.1 & 79.2 & 90.5  & 78.9 & \winAllwoPose{91.1} & \winNoPose{80.0} & 90.7 & \winAll{82.6}  \\
% \hline  
\shline
\end{tabular}
\begin{tablenotes} 
    \item[] $*$ denotes symmetric objects; 
    \item[] $^\dagger$ The best label-free method is marked in bold, and the overall best method is underlined. 
\end{tablenotes}
\end{threeparttable}
\end{table*}

%% file: tables/cropped_lm.tex
\begin{table}
\centering
\caption{\label{tab:lm_crop}
    Results on Cropped LINEMOD
}
\begin{threeparttable}[c]
\tablestyle{10pt}{1.3}
\begin{tabular}{@{}l x{60} x{60} @{}}
% \hline
\shline
Method & \tabincell{c}{Classification\\Accuracy (\%)} & \tabincell{c}{Mean Angle\\Error (\degree)} \\
% \shline
\hline
PixelDA~\cite{bousmalis2017unsupervisedPixelda} & 99.9 & 23.5 \\ 
DRIT~\cite{lee2018diverse}                      & 98.1 & 34.4 \\
DeceptionNet~\cite{zakharov2019deceptionnet}    & 95.8 & 51.9 \\
$\pselfvo$~\cite{wang2020self6d}                & 100.0 & 15.8 \\
% \hline
$\oursLB$                                       & 100.0 & 11.2 \\
$\ours$  -- RGB                                  & 100.0 & 4.7 \\
$\ours$  -- RGB-D                                 & 100.0 & 3.9 \\
% \hline
\shline
\end{tabular}
\end{threeparttable}
\end{table}

%% file: sections/5_conclusion.tex
\section{Conclusion}

In this work, we have introduced $\pself$, the first self-supervised 6D object pose estimation approach aimed at learning from real data without the need for 6D pose annotations. 
Leveraging noisy student training and differentiable rendering, we are able to enforce several visual and geometrical constraints.
In addition, we proposed an occlusion-aware pose estimator to make the self-supervision more robust to challenging scenarios, exploiting both visible and amodal mask information.
Moreover, compared to \cite{wang2020self6d}, we do not naturally depend on depth data during self-supervised training, thanks to noisy-student training and the capabilities of the RGB-based deep refiner~\cite{li2019deepim,labbe2020cosypose}.
To summarize, our method has demonstrated to remarkably reduce the gap towards the state-of-the-art for pose estimation relying on real 6D pose labels.

As future work, it would be very interesting to investigate if our self-supervision can be even applied for unseen objects or categories when no appropriate 3D CAD model is available.
Another interesting aspect is to incorporate also 2D detection into our self-supervision, as this allows backpropagating the loss in an end-to-end fashion throughout both networks including Yolov4.
\Rthree{Furthermore, the development of more lightweight and efficient self-supervised learning methods could be a very interesting and meaningful future direction.}